\documentclass[journal]{IEEEtran}

\usepackage{cite}
\usepackage{soul}
\usepackage{lmodern}
\usepackage{graphicx}
\usepackage{multirow}
\usepackage{times}
\usepackage{ragged2e}
\usepackage{amsmath,amssymb, amsfonts}
\usepackage{tabularx}
\usepackage{hyperref}
\usepackage{cleveref}
\usepackage{siunitx}
\usepackage{dblfloatfix}
\usepackage{authblk}
\usepackage[table]{xcolor}
\usepackage[T1]{fontenc}
\usepackage{siunitx}
\usepackage{booktabs}
\usepackage[caption=false]{subfig}
\usepackage{algorithm}
\usepackage{algpseudocode}
\usepackage{adjustbox}
\usepackage{tabularx}
\usepackage{booktabs}
\usepackage[switch]{lineno}

\begin{document}
\noindent

\title{\LARGE \bf
Model Validity in Observers: When to Increase the Complexity of Your Model?}

\author{
Agapius Bou Ghosn$^{1}$,
Philip Polack$^{1}$,
and Arnaud de La Fortelle$^{1,2}$
\thanks{$^{1}$ Center for Robotics, Mines Paris, PSL University, 75006 Paris, France {\tt [agapius.bou\textunderscore ghosn, philip.polack, arnaud.de\textunderscore la\textunderscore fortelle]@minesparis.psl.eu}}
\thanks{$^{2}$ Heex Technologies, Paris, France}
}

\maketitle

\thispagestyle{empty}
\pagestyle{empty}
\begin{abstract}
Model validity is key to the accurate and safe behavior of autonomous vehicles. Using invalid vehicle models in the different plan and control vehicle frameworks puts the stability of the vehicle, and thus its safety at stake. In this work, we analyze the validity of several popular vehicle models used in the literature with respect to a real vehicle and we prove that serious accuracy issues are encountered beyond a specific lateral acceleration point. We set a clear lateral acceleration domain in which the used models are an accurate representation of the behavior of the vehicle. We then target the necessity of using learned methods to model the vehicle's behavior. The effects of model validity on state observers are investigated. The performance of model-based observers is compared to learning-based ones. Overall, the presented work emphasizes the validity of vehicle models and presents clear operational domains in which models could be used safely. 

\end{abstract}


\section{Introduction}
\IEEEPARstart{T}{he} ability to accurately describe the dynamic behavior of a vehicle is fundamental to reaching full autonomy. Throughout the last decades, various vehicle and tire models have been introduced attempting to represent the vehicle behavior. These constituted the basis of the different control-aimed applications including state estimation, trajectory planning, and control. The validity of the chosen models in these applications is a major factor in their usability. Model validity refers to the adequacy of a specific model in representing the real-world system. Observing the vehicle state, planning vehicle trajectories, or controlling the vehicle using invalid vehicle models leads to inaccurate vehicle behavior. Model validity can be easily ensured when in low excitation drives as the vehicle behavior is almost linear; challenges arise in high excitation drives when nonlinearities dominate the behavior of the vehicle leading to inconsistent vehicle modeling.   
    
The importance of valid vehicle models has been implicitly explored in different observing, planning, and control works. 

In the field of vehicle state estimation, the literature involved introducing modifications to linear tire models (e.g. \cite{reina_vehicle_2019, van_aalst_adaptive_2018}), employing complex vehicle and tire models (e.g. \cite{katriniok_adaptive_2016, wang_estimation_2010}) or estimating the additional road parameters, as the road slope or bank angles (e.g. \cite{zhao_linhui_vehicle_2008, guo_modular_2016, yin_design_2017}); seeking to ensure precise access to the vehicle state.

\begin{figure}[h]
    \centering
    \includegraphics[width=\columnwidth]{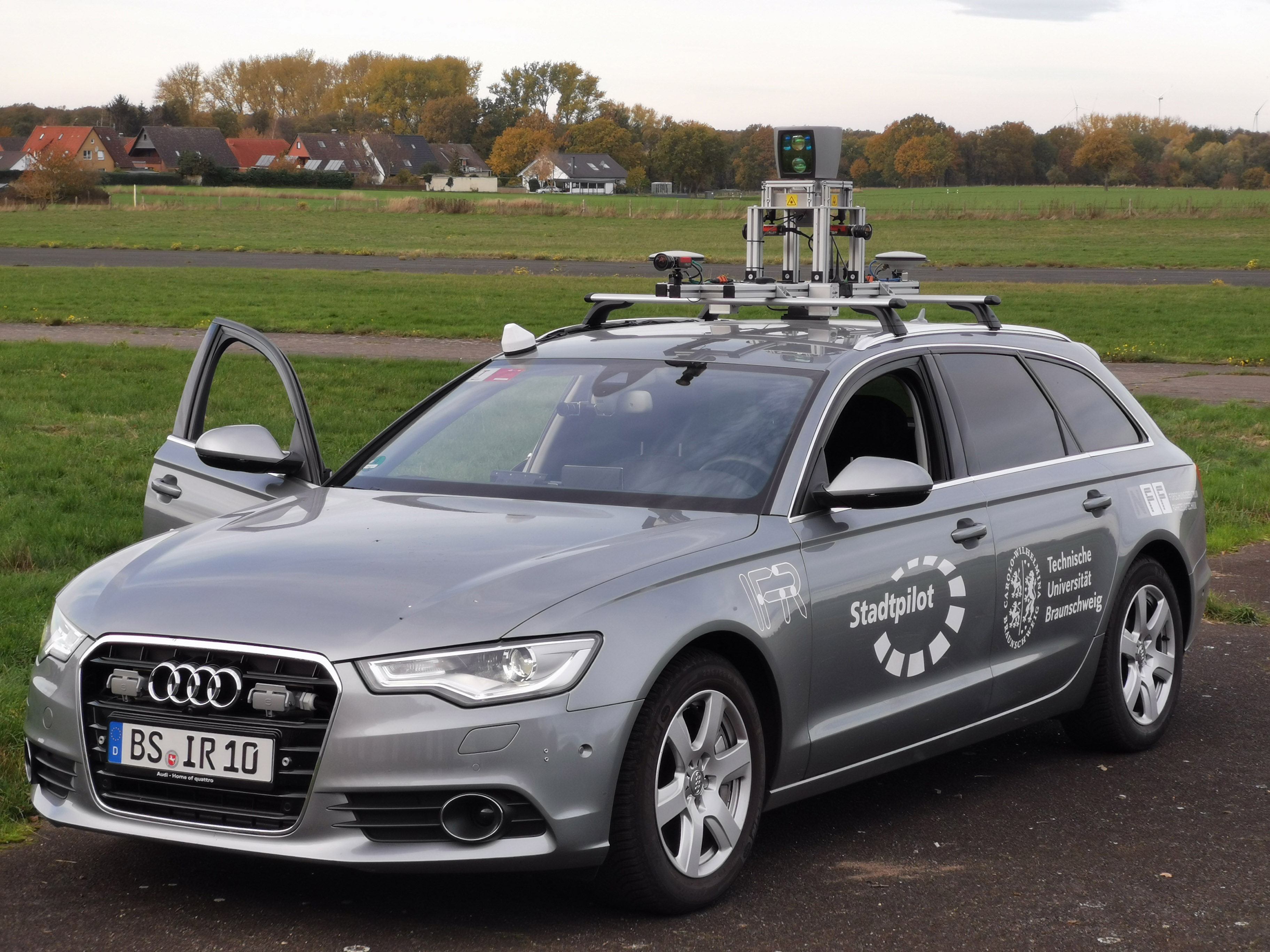}
    \caption{The used experimental vehicle.}
    \label{stadtpilot.fig}
\end{figure}

In the field of trajectory planning, the literature explored ways to ensure valid planner models that lead to feasible trajectory generation. These included setting envelopes and limitations to simple vehicle models (e.g. \cite{polack_kinematic_2017, altche_simple_2017}) to ensure that the resulting models are an accurate representation of the behavior of the vehicle.

In control applications, the literature included adding limits and constraints to used models in model-predictive controllers (e.g. \cite{beal_model_2013, jalali_model_2017}) or including control corrections (e.g. \cite{choi_model_2014}) to ensure a controller able to deal with model errors and to cope with the varying vehicle dynamics. 

Other works have employed learning-based approaches to learn the vehicle dynamics for accurate state estimation (e.g. \cite{revach_kalmannet_2021, choi_split-kalmannet_2023, bou_ghosn_learning-based_2022}), or control (e.g. \cite{williams_aggressive_2016,spielberg_neural_2022}) attempting to reduce the errors between the used model and the actual vehicle system using neural networks. 

An explicit analysis of the validity of different models guarantees their consistent use in autonomous vehicle applications. This work aims to compare the accuracy of different vehicle models to infer their validity, and investigate the model error effects on the accuracy of state observers. For this purpose, four popular vehicle and tire model combinations will be implemented: the dynamic bicycle model with a linear tire model, the dynamic bicycle model with a Dugoff tire model, the dynamic bicycle model with a Pacejka tire model, and the four-wheel model with a Pacejka tire model. Their ability to represent the behavior of the real vehicle shown in Fig. \ref{stadtpilot.fig} will be analyzed. The analyzed models will be then used in Extended Kalman Filters to estimate the longitudinal and lateral velocities and yaw rate of the real vehicle; their accuracy will be analyzed, linked to the model errors, and compared to a learned approach to emphasize the advantages of learned vehicle models over deterministic ones. The contributions of this work are the following:
\begin{itemize}
    \item Analyzing the validity of vehicle models used in the literature in representing real vehicle behavior for different driving scenarios.
    \item Showing clear acceleration limits that affect the ability of models to represent vehicle dynamics.
    \item Showing the effects of model validity on the performance of vehicle state observers. 
    \item Comparing deterministic observing methods to learning-based ones to show their advantages. 
\end{itemize}

In what follows, the model validity problem will be presented in Section \ref{modelValidity.sec}, the state-of-the-art vehicle and tire models will be detailed in Section \ref{soa.sec}, the used experimental vehicle and the data collection procedure will be presented in Section \ref{realData.sec}, the ability of the models to represent the behavior of the real vehicle will be presented in Section \ref{realAnalysis.sec} and the effects on state observing will be presented in Section \ref{realObserving.sec}; the work is concluded in Section \ref{conclusion.sec}.

\section{The Model Validity Problem}\label{modelValidity.sec}
Employing a valid model in vehicle plan and control applications leads to accurate vehicle behavior. Model validity is related to different factors including road conditions and the type of maneuver the vehicle is performing. 

Over the years, many literature works linked the validity of vehicle models to the vehicle's lateral acceleration. In \cite{mitschke_dynamik_1990}, the use of the dynamic bicycle model with a linear tire model was restricted to lateral accelerations $|a_y|<0.4g$, $g$ being the gravitational acceleration. While in \cite{smith_effects_1995}, the same model was assumed to be valid for lateral accelerations $|a_y|<0.2g$ only, with a Dugoff \cite{dugoff_tire_1969} tire model being used for scenarios where $|a_y|>0.5g$. In \cite{polack_kinematic_2017}, the kinematic bicycle model was shown to be valid as long as the lateral acceleration $|a_y|<0.5\mu g$, $\mu$ being the road friction coefficient.

Other works presented comparisons between different vehicle models as in \mbox{\cite{kong_kinematic_2015, kang_comparative_2014}} where the kinematic and the dynamic bicycle with linear tire models were used to estimate the state of vehicle, pointing out to the errors of the dynamic bicycle model with a linear tire model and to the scenarios in which the kinematic bicycle model might be a better representation of the state evolution of the vehicle. In \mbox{\cite{shim_understanding_2007}}, the authors evaluated the validity of simplifying a 14-DoF model to represent a vehicle's roll dynamics; several assumptions were tested, and a comparison with an 8-DoF model was performed to conclude the possible simplifications with minimal effects on the accuracy of the roll representation.

This work will clearly compare the behavior of four popular linear and nonlinear vehicle and tire models in representing a real experimental vehicle's behavior to provide an overview of their performance while focusing on the longitudinal and lateral velocities and yaw rate states of the vehicle. The lateral acceleration conditions will be revisited and conclusions will be made about the effect of using each of the models in different acceleration domains. The effects of using these models in observing applications will be then presented and compared to learned observers.

The models used are defined next.

\section{State-of-the-art Vehicle Models}\label{soa.sec}
As we aim to compare the ability of vehicle models to represent the dynamics of the vehicle, the considered models are introduced in this section. This section will introduce two vehicle models: the four-wheel model and the bicycle model; before moving to three wheel models: the linear model, the Dugoff model and the Pacejka \cite{pacejka_magic_1997} model. Afterwards, the bicycle model will be used with the linear tire model, the Dugoff tire model and the Pacejka tire model while the four-wheel vehicle model will be used with the Pacejka tire model. 

\subsection{Four-wheel Model}
The four-wheel model is a representation of a vehicle's chassis with two front steerable wheels and two rear non-steerable wheels. The motion of the model is represented by the velocity vector issued from the center of gravity ($cog$) of the vehicle. 
It is assumed that the aerodynamic forces are represented by a single force at the vehicle's front. The model considers the coupling of longitudinal and lateral slips and the load transfer between tires. The states, parameters and control inputs describing the model are presented in Table \ref{fourWheelParameters.tab}.

\begin{figure}[h]
    \centering
    \includegraphics[width=\columnwidth]{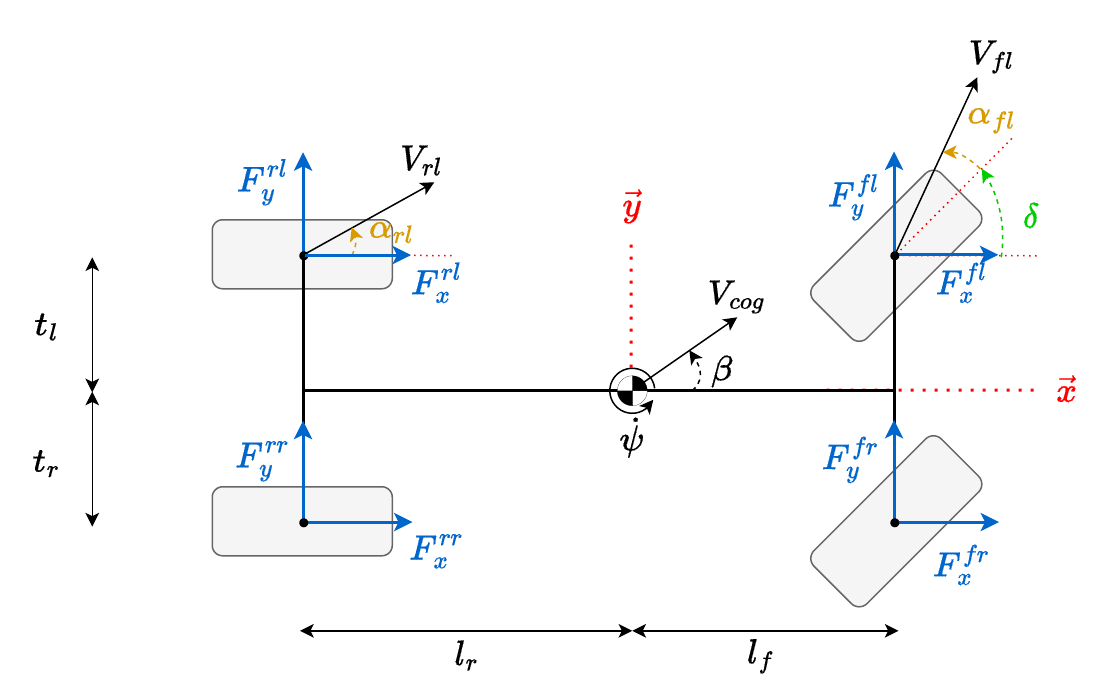}
    \caption{Four-wheel vehicle model.}
    \label{fig:enter-label}
\end{figure}

\begin{table}
  \centering
  \caption{Parameters of the four-wheel model.}
    \begin{tabularx}{\columnwidth}{cXc}
      \toprule
      \textbf{Constants} & \textbf{Characteristics} &\textbf{Unit}\\
      \midrule
      $M_T$ & Total mass of the vehicle & \SI{}{\kilo\gram}\\
      $M_S$ & Suspended mass of the vehicle & \SI{}{\kilo\gram} \\
      $I_x$, $I_y$, $I_z$ & Inertia of the vehicle around its roll, pitch and yaw axis & \SI{}{\kilo\gram\meter\squared}\\
      $l_f$ & Length from the center of gravity of the vehicle to the front wheels & \SI{}{\meter}\\
      $l_r$ & Length from the center of gravity of the vehicle to the rear wheels & \SI{}{\meter}\\
      $t_l$ & Length from the center of gravity of the vehicle to the left wheels & \SI{}{\meter}\\
      $t_r$ & Length from the center of gravity of the vehicle to the right wheels & \SI{}{\meter}\\
      $h_{cog}$ & Height of the center of gravity & \SI{}{\meter}\\
      $h_{aero}$ & Height at which the aerodynamic drag force is applied & \SI{}{\meter}\\
      $r_{eff}$ & Effective tire radius (i.e. distance between the road and the wheel center) & \SI{}{\meter} \\
      $I_r$ & Inertia of the wheel  & \SI{}{\kilo\gram\meter\squared}\\
      \midrule
      \textbf{Variables} & \textbf{Characteristics} &\textbf{Unit}\\
      \midrule
      $\Theta_s$, $\Theta_b$ & Slope and bank angles of the road & \SI{}{\radian} \\
      $T_i$ & Torque applied at wheel $i$ & \SI{}{\newton\meter}\\
      $\delta$ & Angle of the front wheel with respect to the axis of the vehicle (steering angle) & \SI{}{\radian}\\
      $\beta$ & Angle between the velocity vector at the center of gravity and the vehicle's longitudinal axis (side-slip angle) & \SI{}{\radian}\\
      $\alpha_i$ & Slip angle at wheel $i$ & \SI{}{\radian}\\
      $\theta$, $\phi$, $\psi$ & Roll, pitch and yaw angles & \SI{}{\radian}\\
      $V_x$, $V_y$, $V_z$ & Longitudinal, lateral and vertical velocities of the center of gravity in the vehicle frame & \SI{}{\meter\per\second}\\
      $\omega_i$ & Angular speed of wheel $i$ & \SI{}{\radian\per\second}\\
      $V_{xp}^{i}$, $V_{yp}^{i}$ & Longitudinal and lateral velocities of wheel $i$ & \SI{}{\meter\per\second}\\
      $X$, $Y$& Vehicle coordinates in the inertial frame& m\\
      
      \midrule
      \textbf{Forces} & \textbf{Characteristics} &\textbf{Unit}\\
      \midrule
      $F_{x}^{i}$, $F_{y}^{i}$ & Longitudinal and lateral forces applied on wheel $i$ expressed in the vehicle's frame & \SI{}{\newton}\\
      $F_{xp}^{i}$, $F_{yp}^{i}$ & Longitudinal and lateral forces applied on wheel $i$ expressed in the tire's frame & N\\
      $F_{z}^{i}$ & Vertical force at wheel $i$ & \SI{}{\newton}\\
      $F_{s}^{i}$ & Suspension force at wheel $i$ & \SI{}{\newton}\\
      $F_{aero}$ & Aerodynamic force applied to the vehicle & \SI{}{\newton}\\
      \bottomrule
    \end{tabularx}
\label{fourWheelParameters.tab}
\end{table}
The notations employed when representing the vehicle model are presented in Table \ref{fourWheelNotations.tab}.

\begin{table}
  \centering
  \caption{Notations of the four-wheel model.}
  \begin{adjustbox}{width=.8\columnwidth}
    \begin{tabular}{cp{.5\columnwidth}}
      \toprule
      \textbf{Notation} & \textbf{Description}\\
      \midrule
      $cog$ & Center of gravity \\
      $aero$ & Aerodynamic \\
      $fl$ & Front left \\
      $fr$ & Front right \\
      $rl$ & Rear left \\
      $rr$ & Rear right \\
    \bottomrule
    \end{tabular}
     \end{adjustbox}    

\label{fourWheelNotations.tab}
\end{table}

The model is assumed to be on a road with a slope angle $\Theta_s$ and a bank angle $\Theta_b$. The aerodynamic forces are represented by a single force applied at a height $h_{aero}$ with a magnitude $F_{aero} = \frac{1}{2}\rho C_dA_F(V_x+V_{\text{wind}})^2$ with $\rho$ being the mass density of the air, $C_d$ the aerodynamic drag coefficient, $A_F$ the frontal area of the vehicle, $V_x$ the longitudinal vehicle's velocity and $V_{\text{wind}}$ the wind's velocity.

The state of the model is defined as 
$
\begingroup
\setlength\arraycolsep{4.25pt}
Z = \left[\begin{matrix}X & V_x & Y & V_y & \psi & \dot\psi & \theta & \dot\theta & \phi  \end{matrix}\begin{matrix}& \dot\phi
&\omega_1 & \omega_2 & \omega_3 & \omega_4\end{matrix}\right]^T
\endgroup
$
and its state evolution is described by the following equations: \begin{subequations}\label{fourWheelTranslation.eq}
\begin{eqnarray}
M_T \dot{V_x} &=& M_T \dot\psi V_y + (F_{x}^{fl}+F_{x}^{fr}+F_{x}^{rl}+F_{x}^{rr}) + \label{4wm1.eq}\\
&& M_Tg\sin(\phi-\Theta_s) - F_{aero}\cos\phi \nonumber\\
M_T \dot{V_y} &=& -M_T \dot\psi V_x + (F_{y}^{fl}+F_{y}^{fr}+F_{y}^{rl}+F_{y}^{rr})- \label{4wm2.eq}\\
&&  M_Tg\sin(\theta-\Theta_b)\cos(\phi-\Theta_s) \nonumber\\
M_S \dot{V_z} &=& M_S \dot\phi V_x + (F_{z}^{fl} + F_{z}^{fr} + F_{z}^{rl} + F_{z}^{rr}) -\\
&& M_Sg\cos(\theta-\Theta_b)\cos(\phi-\Theta_s) + F_{aero}\sin\phi \nonumber
\end{eqnarray}
\end{subequations}
The evolution of the coordinates of the vehicle in the inertial frame are defined by the following equations:
\begin{subequations}\label{xyLink.eq}
\begin{eqnarray}
\dot X &=& V_x\cos\psi-V_y\sin\psi \\
\dot Y &=& V_x\sin\psi+V_y\cos\psi 
\end{eqnarray}\label{inertialConversion.eq}
\end{subequations}
The evolution of the roll, pitch and yaw angular velocities are defined as:
\begin{subequations}\label{fourWheelRotation.eq}
\begin{eqnarray}
I_x \ddot\theta &=&-(F_{s}^{fr}+F_{s}^{rr})t_r+(F_{s}^{fl}+F_{s}^{rl})t_l \\
&& + (F_{y}^{fl}+F_{y}^{fr}+F_{y}^{rl}+F_{y}^{rr})h_{cog} \nonumber\\
I_y \ddot\phi &=& -(F_{s}^{fl}+F_{s}^{fr})l_f + (F_{s}^{rl}+F_{s}^{rr})l_r -  \\
&&(F_{x}^{fl}+F_{x}^{fr}+F_{x}^{rl}+F_{x}^{rr})h_{cog} + \nonumber \\ 
&& (h_{aero}-h_{cog})F_{aero} \nonumber\\
I_z \ddot\psi &=& (F_{y}^{fl} + F_{y}^{fr})l_f - (F_{y}^{rl}+F_{y}^{rr})l_r + \label{4wm3.eq}\\*
&&(F_{x}^{fr}+F_{x}^{rr})t_r-(F_{x}^{fl}+F_{x}^{rl})t_l \nonumber
\end{eqnarray}
\end{subequations}

The longitudinal $F_{xpi}$ and lateral $F_{ypi}$ tire forces are functions of each wheel's slip ratio $\tau_{xi}$, side slip angle $\alpha_i$, reactive normal force $F_{zi}$ and the friction coefficient of the road $\mu$:
\begin{subequations}
\begin{eqnarray}
F_{xpi} &=& f_x(\tau_{xi}, \alpha_i, F_{zi}, \mu)\\
F_{ypi} &=& f_y(\tau_{xi}, \alpha_i, F_{zi}, \mu)
\end{eqnarray}
\end{subequations}
$f_x$ and $f_y$ being defined by the tire model. The slip ratio and slip angle being defined by the following equations:
\begin{subequations}
\begin{equation}\label{slip.eq}
\tau_{xi} = \begin{cases} \frac{r_{eff}\omega_i-V_{xpi}}{r_{eff}\omega_i} & \text{if $r_{eff}\omega_i \geq V_{xpi}$} \\
\frac{r_{eff}\omega_i-V_{xpi}}{|V_{xpi}|} & \text{if $r_{eff}\omega_i<V_{xpi}$} 
\end{cases}
\end{equation}
\begin{eqnarray}
\alpha_f &=& \delta_f-\arctan \left(\frac{V_y+l_f\dot\psi}{V_x\pm l_w\dot\psi}\right) \\
\alpha_r &=& -\arctan \left(\frac{V_y-l_r\dot\psi}{V_x\pm l_w\dot\psi}\right) 
\end{eqnarray}
\end{subequations}
The conversion between the tire forces in the tire frame and the tire forces in the vehicle frame follows the following equations: 
\begin{subequations}\label{tirebodylink.eq}
\begin{eqnarray}
F_{xi} &=& (F_{xpi}\cos\delta_i-F_{ypi}\sin\delta_i)\cos\phi-F_{zi}\sin\phi \\
F_{yi} &=& (F_{xpi}\cos\delta_i-F_{ypi}\sin\delta_i)\sin\theta\sin\phi+ \\
&& (F_{ypi}\cos\delta_i+F_{xpi}\sin\delta_i)\cos\theta+F_{zi}\sin\theta\cos\phi \nonumber
\end{eqnarray}
\end{subequations}

The vertical and tire forces will be introduced later on. The bicycle model is introduced next.

\subsection{Bicycle Model}
\begin{figure}[h]
    \centering
    \includegraphics[width=\columnwidth]{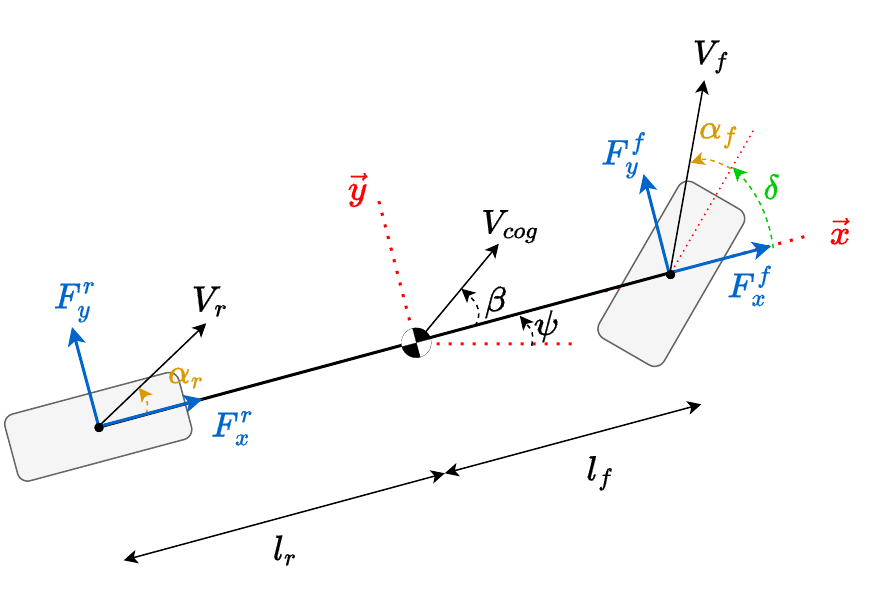}
    \caption{The dynamic bicycle model.}
    \label{bicycleModel.fig}
\end{figure}

The bicycle model is a simplification of the previously introduced four-wheel model. The bicycle model is shown in Fig. \ref{bicycleModel.fig}. The model assumes that the  four-wheel model can be lumped into a bicycle model: the two front wheels are represented by a single steerable front wheel and the two rear wheels are represented by a single non-steerable rear wheel; the road bank angle is neglected; the pitch, roll and vertical dynamics are neglected.

The state of the model is defined as $Z=\begin{bmatrix} X & V_x & Y & V_y & \psi & \dot\psi \end{bmatrix}$ and its evolution is described by the following equations: 
 \begin{subequations}\label{DBM.eq}
     \begin{eqnarray}
         M_T \dot V_x &=& M_T\dot\psi V_y + F_{x}^{f} + F_{x}^{r} \\
         M_T \dot V_y &=& -M_T\dot\psi V_x + F_{y}^{f} + F_{y}^{r} \\
         I_z \ddot\psi &=& F_{y}^{f}l_f - F_{y}^{r}l_r
     \end{eqnarray}
 \end{subequations}
The link between the tire forces and the bicycle body can be made through Equations (\ref{tirebodylink.eq}) while neglecting the roll and pitch angles; while the link to the X and Y coordinates can be made using Equations (\ref{xyLink.eq}). 

Different tire models are defined next.

\subsection{Tire Models}
The equations of motion of the vehicle depend on the tire forces exerted on the vehicle's body. Modeling the tire forces is linked to various tire models. These range from complex tire models as the Pacejka \cite{pacejka_magic_1997} model, to simple ones as the linear tire model. In this work, we present the Pacejka model, the Dugoff model and the linear model. 

\subsubsection{Pacejka Model}
The Pacejka model is a semi-empirical model also known as the Magic Formula. The general form defined in \cite{pacejka_tyre_2006} is the following: 
\begin{subequations}
\begin{eqnarray}
y &=& D\sin(C\arctan(Bx-E(Bx-  \\
&&\arctan(Bx)))) \nonumber\\
Y &=& y+S_v\\
x &=& X+S_h
\end{eqnarray}
\end{subequations}
Y represents the output variable: the longitudinal force, the lateral force or the self-aligning torque; and X represents the longitudinal slip ratio (or slip angle). The parameters as defined in \cite{pacejka_tyre_2006} are presented in Table \ref{pacejkaParameters.tab}.

\begin{table}
  \centering
      \caption{Parameters of the Pacejka model.}
  \begin{adjustbox}{width=\columnwidth}
    \begin{tabular}{cp{.4\columnwidth}c}
      \toprule
      \textbf{Parameter} & \textbf{Definition} &\textbf{Value}\\
      \midrule
      $B$ & Stiffness factor & $\frac{\left (\frac{dy}{dx} \right)_{x=0}}{CD}$ \\ [10pt]
      $C$ & Shape factor & $1\pm(1-\frac{2}{\pi}\arcsin(\frac{y_s}{D}))$ \\ [10pt]
      $D$ & Peak factor & $y_{max}$\\ [10pt]
      $E$ & Curvature factor & $\frac{Bx_m-\tan(\frac{\pi}{2C})}{Bx_m-\arctan(Bx_m)}$ \\ [10pt]
      $S_h$ & Horizontal shift & - \\ [10pt]
      $S_v$ & Vertical shift & - \\ [10pt]
      $y_s$ & Asymptotic value of $y$ & - \\ [10pt]
      $x_m$ & $x$ value corresponding to $y_{max}$ & - \\ [10pt]
      \bottomrule
    \end{tabular}
     \end{adjustbox}    

\label{pacejkaParameters.tab}
\end{table}

The Pacejka model is known for its accurate tire forces representation but requires many parameters to be able to function. The full set of equations and parameters can be found at the reference. 

\subsubsection{Dugoff Model}
The Dugoff tire model is an analytical tire model introduced in \cite{dugoff_tire_1969}. In comparison with the Pacejka model, it is able to model the tire forces using fewer parameters. It describes the tire forces using the following equations:
\begin{subequations}\label{dugoff.eq}
\begin{eqnarray}
    F_{xp} &=& C_\tau \frac{\tau_x}{1-\tau_x} f(\lambda) \\
    F_{yp} &=& C_\alpha \frac{\tan\alpha}{1-\tau} f(\lambda) \\
    \lambda &=& \frac{\mu F_z (1+\tau)}{2 \sqrt{(C_\tau\tau)^2+(C_\alpha\tan\alpha)^2}} \\
    f(\lambda) &=& \begin{cases} (2-\lambda)\lambda & \text{if $\lambda < 1$}\\
                                1 & \text{if $\lambda \geq 1$}
    \end{cases}
\end{eqnarray}
\end{subequations}
with $C_\tau$ and $C_\alpha$ being the longitudinal and lateral tire cornering stiffnesses.
The Dugoff model has lower accuracy than the Pacejka model in high slip scenarios but is able to represent the tire forces in a wide variety of scenarios with a low number of parameters. 

\subsubsection{Linear Model}
The linear tire model presents a linear relationship between the tire forces and the slip ratio and slip angle.  It is described using the following equations: 
\begin{subequations}\label{linear.eq}
    \begin{eqnarray}
        F_x &=& C_\tau \tau_x \\
        F_y &=& C_\alpha \alpha 
    \end{eqnarray}
\end{subequations}
It is valid only for the linear region of the tire behavior. It overestimates the forces for high slips.

\begin{figure}
    \centering
    \includegraphics[width=\columnwidth]{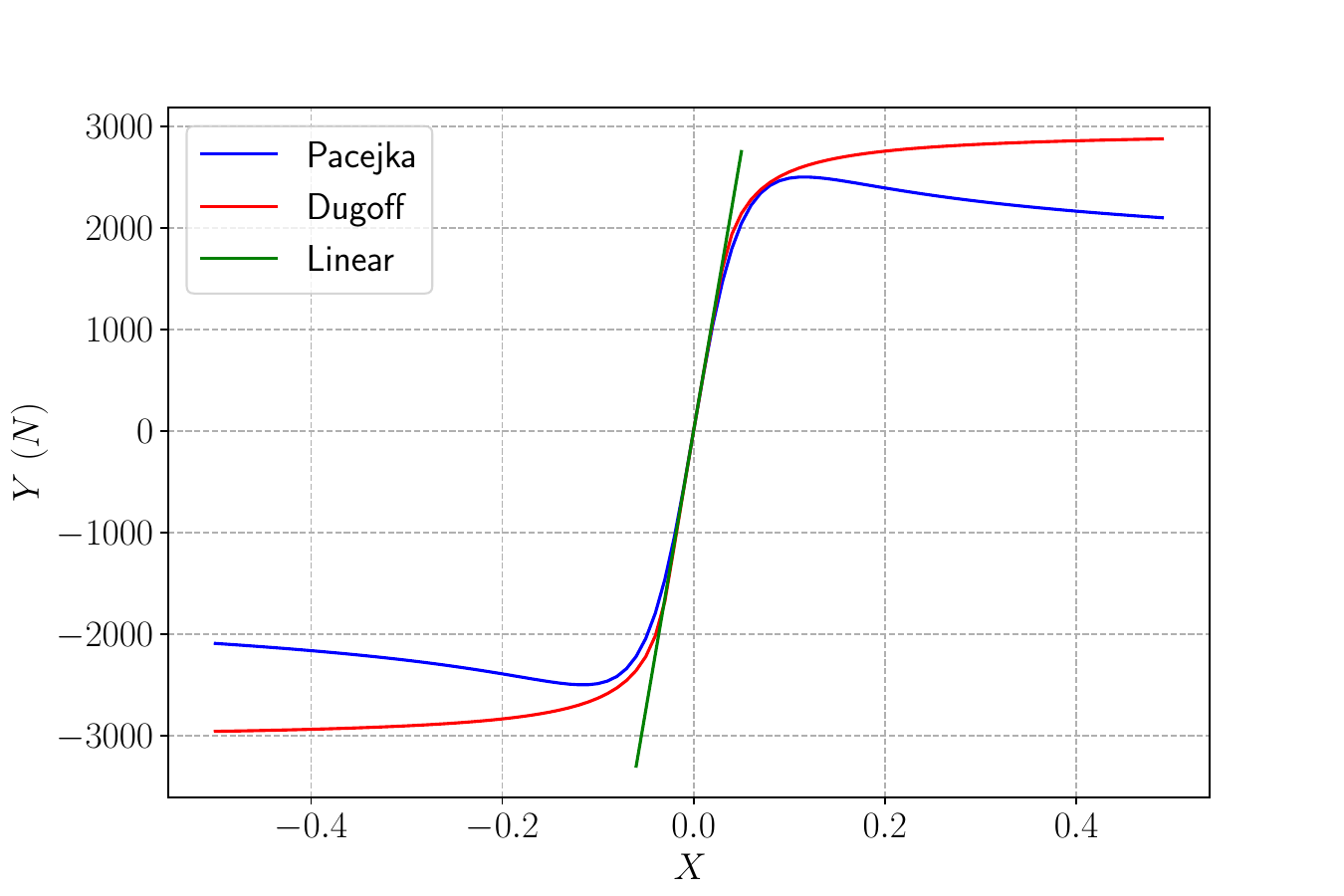}
    \caption{A comparison plot between the different tire models showing the behavior difference with higher slips. (For representation purposes $B=10$, $C=2.2$, $D=2500$, $E=1$, $F_z=3000 \si{\newton}$)}
    \label{tireComp.fig}
\end{figure}

A comparison between the behavior of the three introduced tire models can be seen in Fig. \mbox{\ref{tireComp.fig}}. Taking the Pacejka model as the reference, the figure shows that the linear model is valid only for the linear region of the tire behavior; while the Dugoff model can represent the tire behavior for a larger range even when nonlinearities are available.

Having defined the different vehicle and tire models, the four-wheel model will be used with the Pacejka tire model as done in \mbox{\cite{katriniok_adaptive_2016}}: this implies neglection of the road slope and bank angles, in addition to the pitch and roll dynamics. The dynamic bicycle model will be used with the linear tire model as in \mbox{\cite{kiencke_observation_1997}}, with the Dugoff tire model as in \mbox{\cite{song_pneumatic_2014}} and with the Pacejka tire model as in \mbox{\cite{rucco_dynamics_2010}.} For both the four-wheel model and the dynamic bicycle with the Dugoff and Pacejka tire models, the vertical forces are assumed to be: 

\begin{subequations}\label{Fz.eq}
    \begin{eqnarray}
        F_z^f &=& M \left(\frac{l_r}{l_f+l_r}g-\frac{h_{cog}}{l_f+l_r}a_x\right). \nonumber\\ 
        && \left(\frac{1}{2}\pm \frac{h_{cog}}{(t_l+t_r)g}a_y\right) \\
        F_z^r &=& M \left(\frac{l_f}{l_f+l_r}g+\frac{h_{cog}}{l_f+l_r}a_x\right). \nonumber\\ 
        &&\left(\frac{1}{2}\pm \frac{h_{cog}}{(t_l+t_r)g}a_y\right) 
    \end{eqnarray}
\end{subequations}
with the accelerations:
\begin{subequations}\label{acc.eq}
    \begin{eqnarray}
        a_x &=& \dot V_x - \dot\psi V_y \\
        a_y &=& \dot V_y + \dot\psi V_x
    \end{eqnarray}    
\end{subequations}
For the four specified models, the control inputs are the wheel speeds used to calculate the slip ratio (using Equations (\mbox{\ref{slip.eq}})), and the steering angle.

\subsection{Learned Vehicle Models}
While deterministic vehicle models are prone to modeling errors many of the literature works tend to use learning-based methods, employing neural networks to estimate the dynamics of the vehicle. Learning-based methods are often used in direct vehicle observing \cite{ghosn_robust_2023, bou_ghosn_learning-based_2022, escoriza_data-driven_2021} or controlling \cite{spielberg_neural_2022, williams_information-theoretic_2018} applications. Learned models show the ability to adapt to low and high dynamic drives, to changes in the measurement noise, and to changes in the environment. 

To be able to compare the different models to the behavior of a real experimental vehicle, the used experimental vehicle is presented in the next section and the data collection procedure is described.

\section{Real Vehicle Definition and Data Collection}\label{realData.sec}
\begin{figure}[h]
    \centering
    \includegraphics[width=\columnwidth]{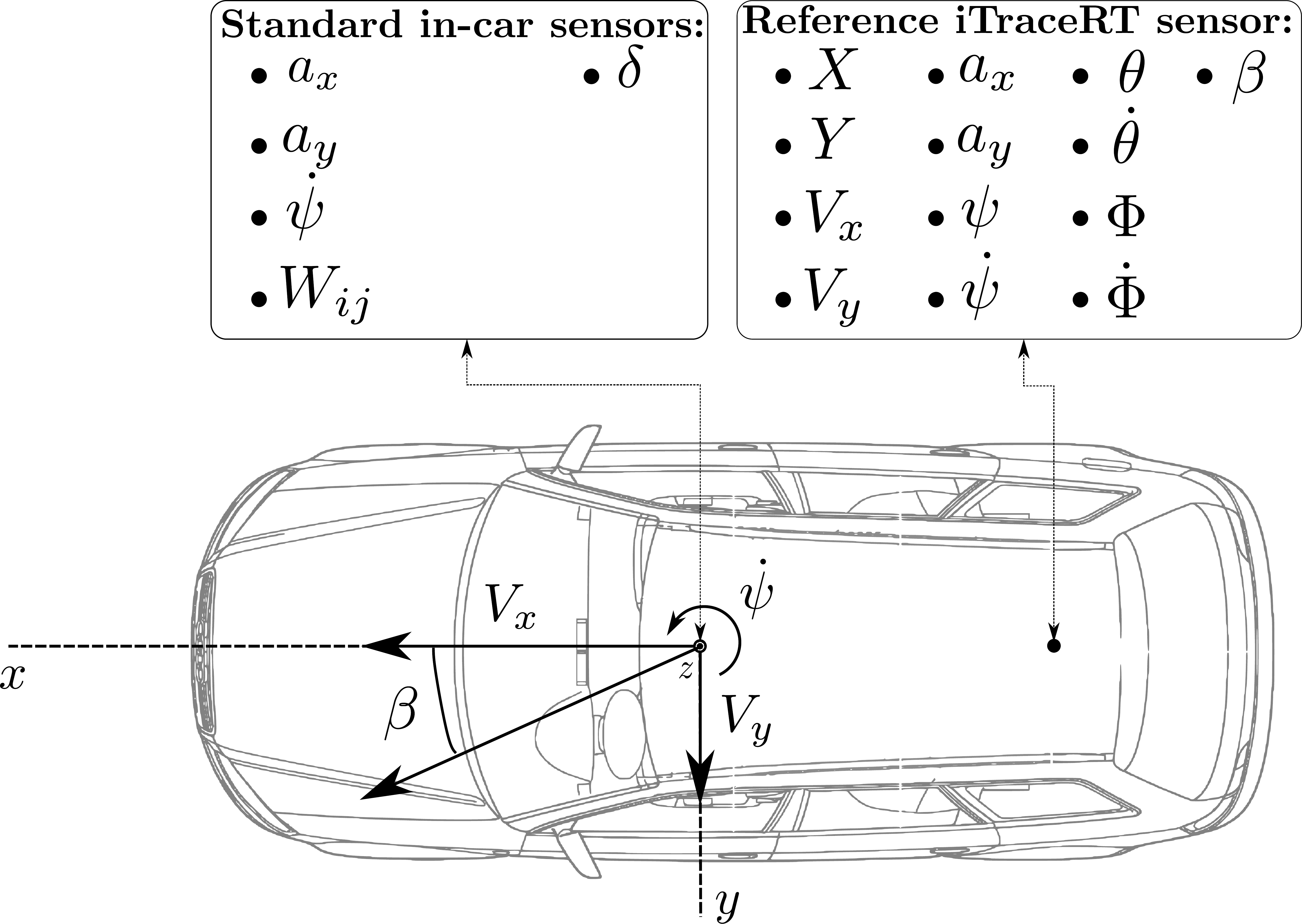}
    \caption{Experimental sensor setup.}
    \label{sensorSetup.fig}
\end{figure}

The used vehicle in our work is the Audi A6 Avant C7 shown in Fig. \ref{stadtpilot.fig}. The characteristics of the vehicle are shown in Table \ref{vehicleChs.tab}.
\begin{table}
\vspace{.2cm}
\centering
    \caption{Parameters of the vehicle used for data collection (CoG: center of gravity)}
    \begin{tabular}{cp{.5\columnwidth}c}
      \toprule
      \textbf{Parameter} & \textbf{Description} & \textbf{Value} \\
      \midrule
      $M_T$ & Mass of the vehicle & \SI{1578}{\kilo\gram}\\
      $l_\mathrm{f}$ & length from CoG to the front axle & \SI{1.134}{\meter} \\
      $l_\mathrm{r}$ & length from CoG to the rear axle  & \SI{1.578}{\meter} \\
      $b$ & Track width & \SI{1.513}{\meter} \\
      $I_z$ & Moment of inertia around the $z$-axis & \SI{2924}{\kilo\gram\square\meter}\\
      \bottomrule
    \end{tabular}
\label{vehicleChs.tab}
\end{table}
The vehicle is equipped with a conventional inertial measurement unit (IMU), the Audi Sensor Array (SARA), which measures the longitudinal and lateral accelerations ($a_x$, $a_y$), the yaw rate ($\dot\psi$), and the wheel speeds ($W_{ij}$). Steering angle measurements ($\delta$) are also available. A reference dual-antenna INS/GNSS \href{https://www.imar-navigation.de/en/products/by-product-names/item/itracert-f200-itracert-f400-itracert-mvt}{iTraceRT F400} sensor provides highly accurate measurement of the position (Easting -- $X$ , Northing -- $Y$) in UTM-coordinates, the longitudinal and lateral velocities ($V_x$, $V_y$), the longitudinal and lateral accelerations ($a_x$, $a_y$), the yaw ($\psi$), pitch ($\theta$), and roll ($\Phi$) angles and rates ($\dot\psi$, $\dot\theta$, $\dot\Phi$), as well as the side-slip angle ($\beta$). It will be considered as the ground truth. The sensor setup of the vehicle is shown in Fig. \ref{sensorSetup.fig}. The iTraceRT sensor provides measurements at $\SI{100}{\hertz}$; the SARA sensors provide measurements at $\SI{50}{\hertz}$. 

The presented vehicle is used for data collection. 

\begin{figure}[h]
    \centering
    \includegraphics[width=.8\columnwidth]{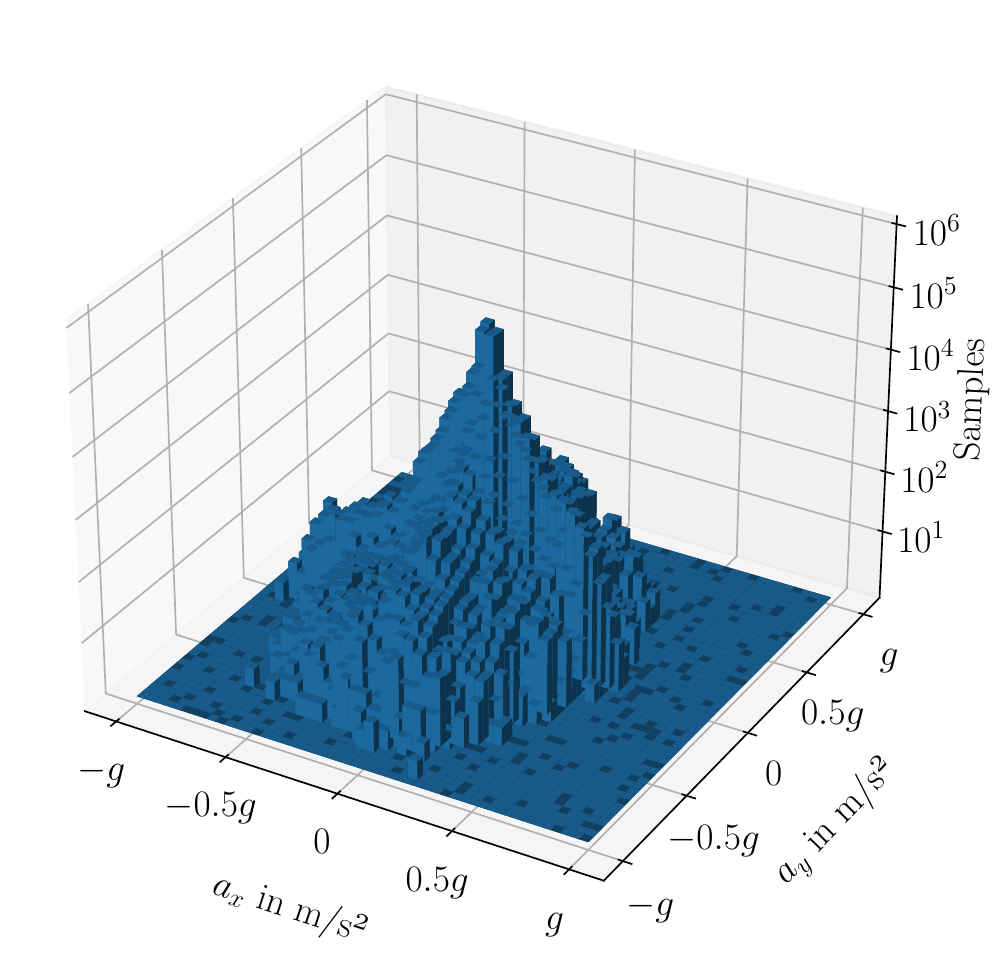}
    \caption{Distribution of the collected samples on the friction circle (samples are shown on a logarithmic axis). The plot shows a data set containing both low and high accelerations with a bias towards lower accelerations.}
    \label{frictionCircle.fig}
\end{figure}

The data collection procedure aims to create a diverse data set containing low and high acceleration maneuvers. The collection procedure involves inner-city maneuvers in Braunschweig, Germany, resulting in low accelerations, and maneuvers effected on a specialized test track near Peine, Germany, resulting in high acceleration maneuvers. The total data set size is 1 million samples. The distribution of the collected samples plotted on the friction circle is presented in Fig. \ref{frictionCircle.fig}. The plot shows that the data set includes both low acceleration and high acceleration maneuvers reaching $|a_y|=1g$.


\section{Real Vehicle Behavior Analysis}\label{realAnalysis.sec}
Having presented the different vehicle models, and collected the needed data sets, we proceed to compare the performance of each of the models with respect to the vehicle. For this purpose, the comparison procedure is presented in Section \ref{comparisonAlg.ssec} and the comparison results are presented in Section \ref{comparisonRes.ssec}. A learning-based method is presented in the state observing application later on.

\subsection{Comparison procedure}\label{comparisonAlg.ssec}
We define the used procedure to compare the four models: the dynamic bicycle model with the linear tire model, the dynamic bicycle model with the Dugoff tire model, the dynamic bicycle model with the Pacejka tire model, and the four-wheel model with the Pacejka tire model. Note that all of the model parameters of the vehicle are provided by the host laboratory and were identified formerly in a separate work \mbox{\cite{wille_stadtpilot_2010}}.   

The considered state variables for comparison are the longitudinal velocity $V_x$, the lateral velocity $V_y$ and the yaw rate $\dot\psi$, constituting the states of the dynamic bicycle models and present in the state vector of the four-wheel model. These variables represent the two-dimensional state evolution of the vehicle. The state to be compared is then defined as $Z = \begin{bmatrix}
        V_x & V_y & \dot\psi
    \end{bmatrix}$. In the comparison performed next, the state evolution equations governing each of the vehicle or tire models presented in Section \ref{soa.sec} are used. A discretization time-step of $\Delta T = \SI{0.02}{\second}$ is imposed by the sensors of the experimental vehicle as shown above.

The comparison algorithm will simply compare the state of the models at each time step, given the previous ground truth state, with the current ground truth state. In other words at each time step, the state of each model is set to the ground truth state, its dynamics equations are then used to compute the next state, and the resulting state is compared to the ground truth data through an absolute error computation. This will eliminate the possibility of error accumulation and will assess the ability of the model to accurately represent the state evolution of the vehicle.  The algorithm used to calculate the state errors $e$ is presented in Algorithm \ref{1.alg} with `gt' being the ground truth values from the iTraceRT vehicle sensor. The inputs to the models are the wheel speeds and steering angle collected from the car sensors. 

\begin{algorithm}
\caption{Comparison Algorithm}\label{1.alg}
\begin{algorithmic}
\State \textbf{Given:}
$f, \Delta T$: Dynamics function, step size;\\
$U$: input;\\
$N$: Data subset size;
\For {$k \gets 0$ \textbf{to} $N$}
    \State $Z^k_{\text{model}} \gets Z^k_{\text{gt}}$;
    \State $Z^{k+1}_{\text{model}} \gets f(Z^{k}_{\text{model}}, U^{k+1}, \Delta T)$;
    \State $e_{k+1} \gets |Z^{k+1}_{\text{model}} - Z^{k+1}_{\text{gt}}|$;
\EndFor
\end{algorithmic}
\end{algorithm}

As seen in the presented algorithm, the three-dimensional error is calculated at each step. These errors will be stored to be analyzed next. 

The presented algorithm is run on a set containing 28 trajectories chosen from the collected data set to depict a change in the performed maximum lateral acceleration as presented in the next section. The tested trajectories constitute a total of 132,000 samples adding up to approximately 45 minutes of driving.

The computed errors will be analyzed next. 

\subsection{Comparison Results}\label{comparisonRes.ssec}
Having presented the model comparison algorithm in the previous section and computed the errors associated with each model, we analyze the error distribution in this section. Throughout the different analysis procedures performed, a clear link was found between the evolution of the errors and the change in the vehicle's lateral accelerations as pointed out in the literature works. For this reason, we start by presenting a visualization of the error evolution with respect to changes in lateral acceleration before proceeding to further tests and conclusions.

\subsubsection{Error visualization}
\begin{figure}[h]
    \centering
    \includegraphics[width=\columnwidth]{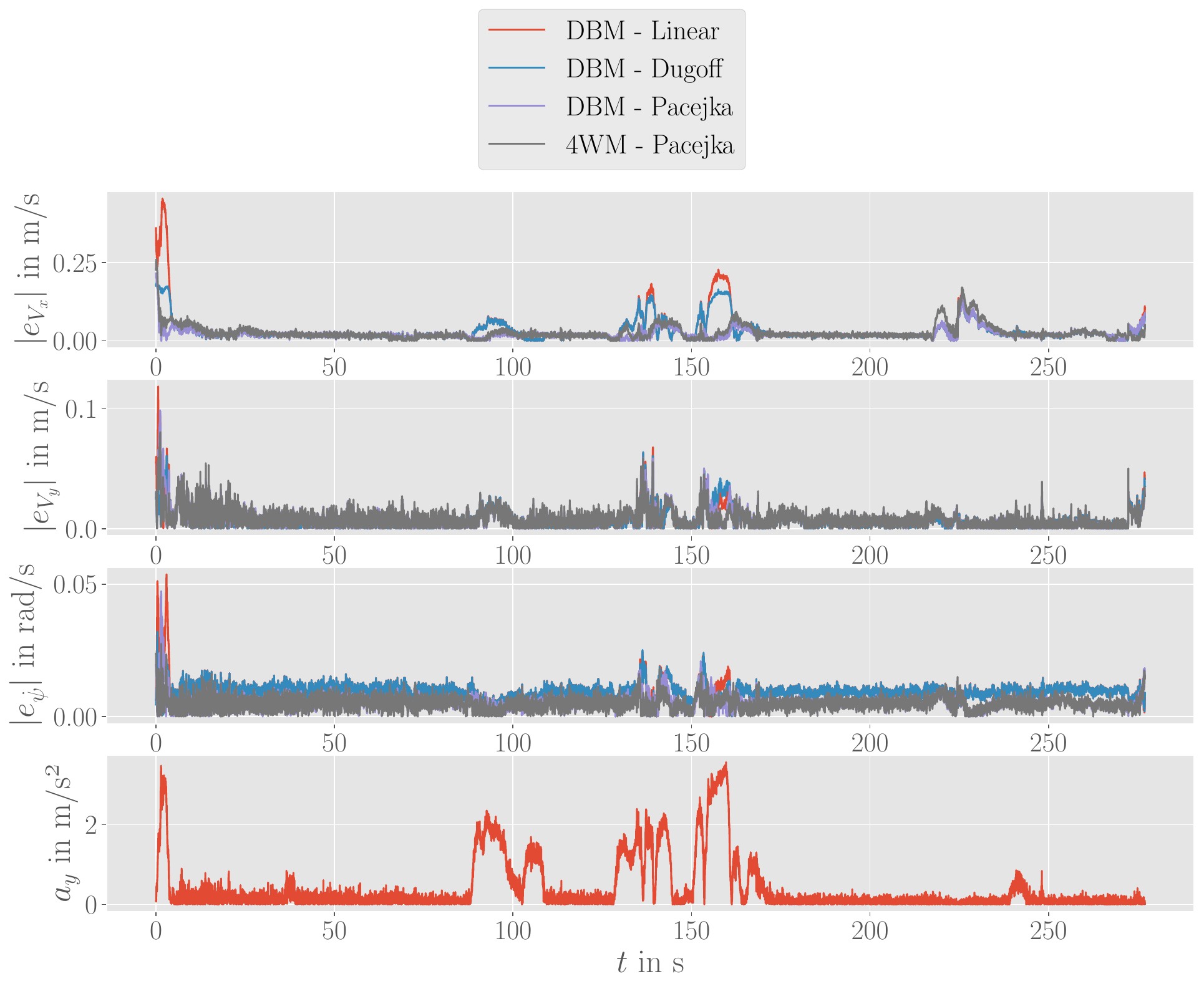}
    \caption{Visualization of the errors of each model in parallel to the lateral acceleration evolution for a low dynamic maneuver ($a_y^{\text{max}} = \SI{3.4}{\meter\per\second\squared}$). Rises in lateral acceleration are often associated with rises in model errors with the linear tire model delivering the highest errors. (DBM: dynamic bicycle model; 4WM: four-wheel model)}
    \label{lowD-vis.fig}
\end{figure}

\begin{figure}[h]
    \centering
    \includegraphics[width=\columnwidth]{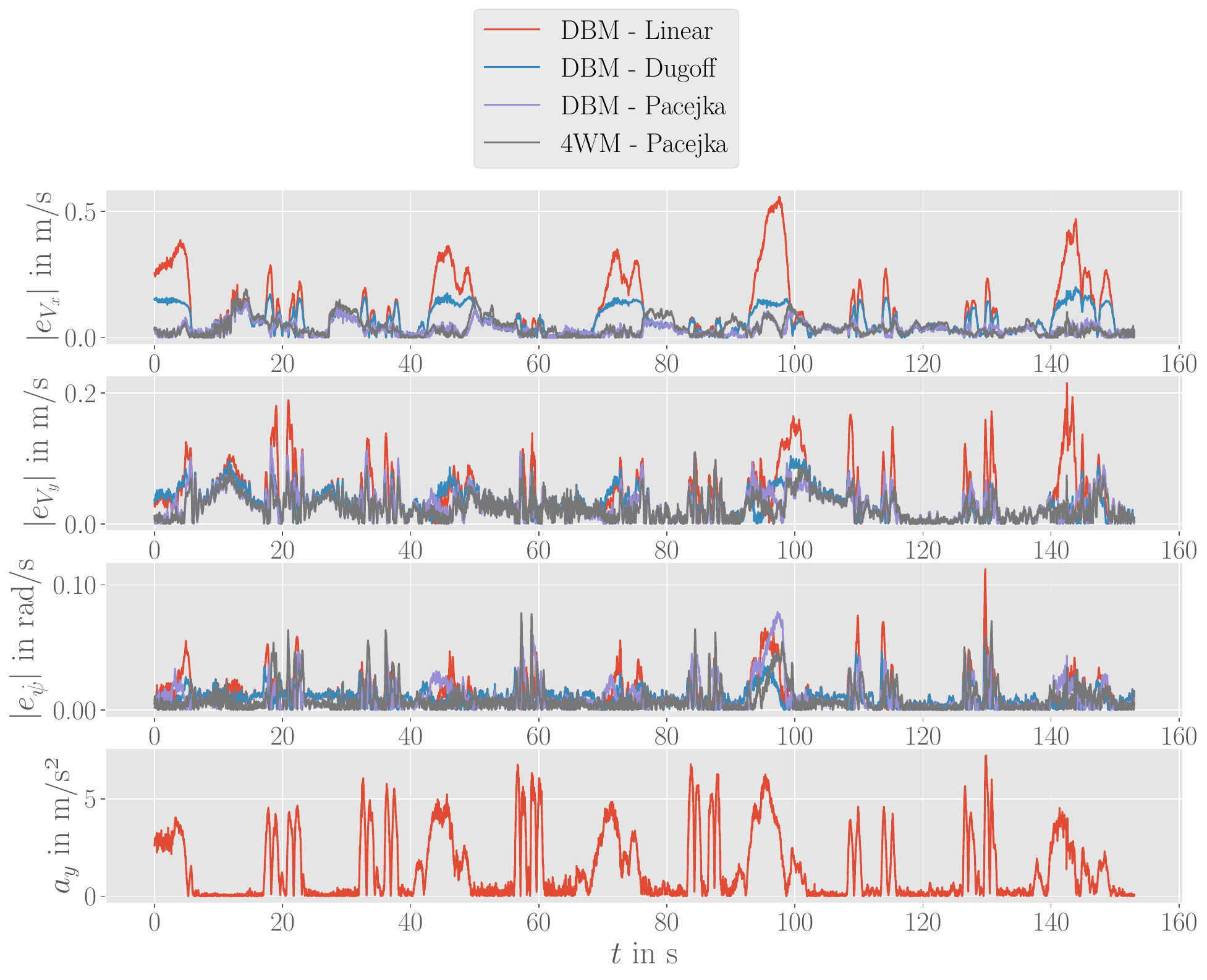}
    \caption{Visualization of the errors of each model in parallel to the lateral acceleration evolution for a higher dynamic maneuver ($a_y^{\text{max}} = \SI{7.0}{\meter\per\second\squared}$). Rises in lateral acceleration are clearly associated with rises in model errors. The linear model shows the highest errors while the Pacejka-based models present a similar behavior. (DBM: dynamic bicycle model; 4WM: four-wheel model)}
    \label{medD-vis.fig}
\end{figure}

\begin{figure}[h]
    \centering
    \includegraphics[width=\columnwidth]{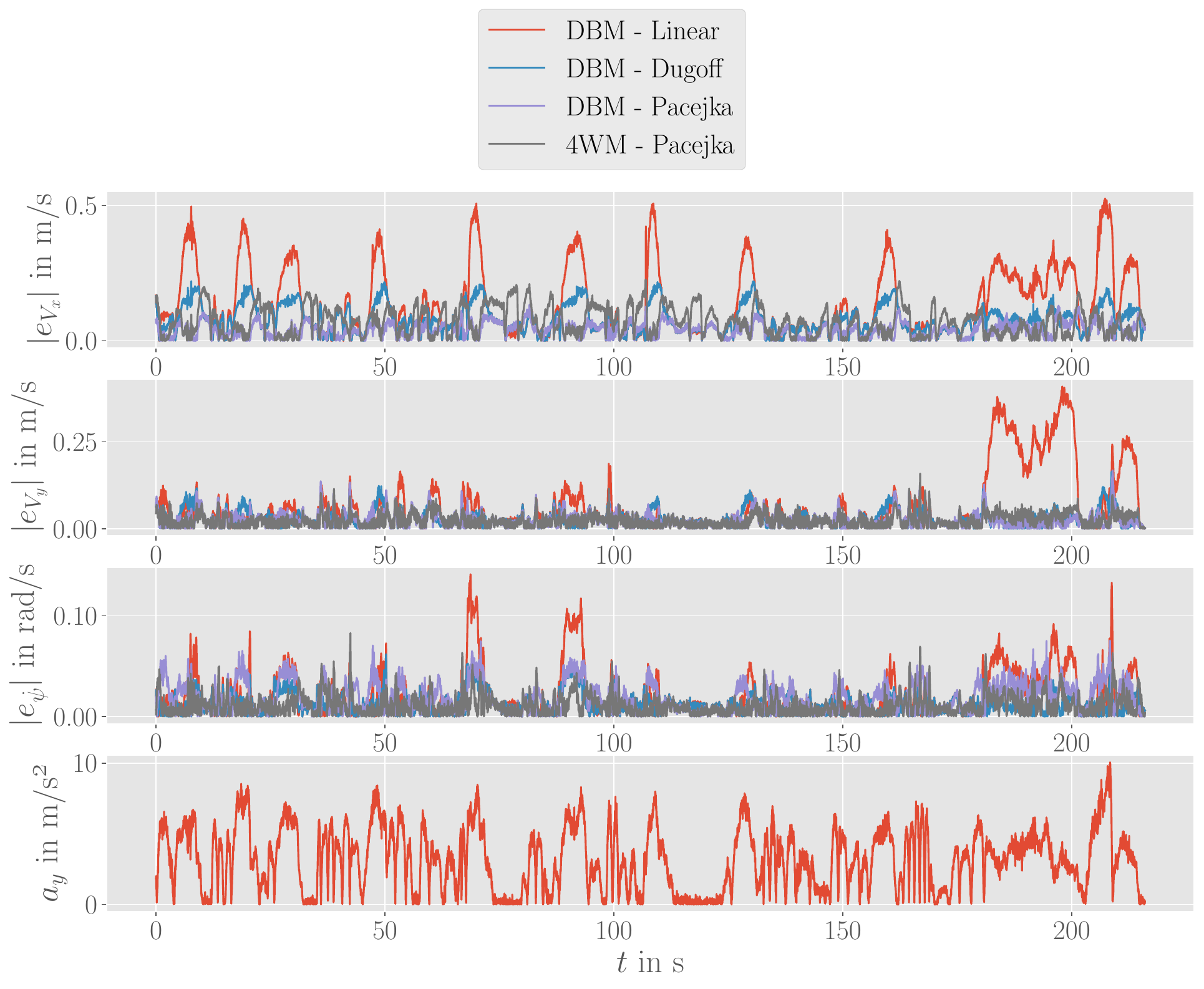}
    \caption{Visualization of the errors of each model in parallel to the lateral acceleration evolution for a near-limits maneuver ($a_y^{\text{max}} = \SI{10.1}{\meter\per\second\squared}$). Rises in lateral acceleration result in rises in model errors. The linear model shows the highest errors. The Pacejka models (bicycle and four-wheel) show the lowest errors for $V_y$ and $\dot\psi$ while their behavior is almost similar to the Dugoff model for $V_x$. (DBM: dynamic bicycle model; 4WM: four-wheel model)}
    \label{highD-vis.fig}
\end{figure}

The error visualization is split into three scenarios depending on the maximum lateral acceleration of the considered trajectory. Three trajectories are considered depicting a low dynamic maneuver with $a_y^{\text{max}} = \SI{3.4}{\meter\per\second\squared}$, a higher dynamic maneuver with $a_y^{\text{max}} = \SI{7.0}{\meter\per\second\squared}$ and a near-limits maneuver with $a_y^{\text{max}} = \SI{10.1}{\meter\per\second\squared}$. The evolution of the absolute errors is plotted in parallel with the evolution of the lateral accelerations throughout the trajectory. Figures \ref{lowD-vis.fig}- \ref{highD-vis.fig} show the three scenarios respectively.

It can be seen in the three plots, that a rise in the lateral accelerations of the vehicle is associated with a rise in the errors of the different models. The higher acceleration maneuvers are associated with higher errors for the four models. The linear-based dynamic bicycle model shows the highest errors even for the low dynamic scenario. The Dugoff model and both Pacejka models show lower errors than the linear model but are as well sensitive to the change in lateral acceleration: this is seen in the two dynamic scenarios. The behavior of both Pacejka-based models is close in the different scenarios and is the most accurate among the other tire models.

It is remarked that the high dynamic scenarios are associated with higher sensitivity to changes in lateral accelerations as the errors in the first scenario remain bounded whereas chaotic performances are seen in the two dynamic scenarios.

To further investigate the relationship between the lateral accelerations and the modeling errors, the whole test set will be analyzed next.



\subsubsection{Error Analysis}
\begin{figure}[h]
    \centering
    \includegraphics[width=\columnwidth]{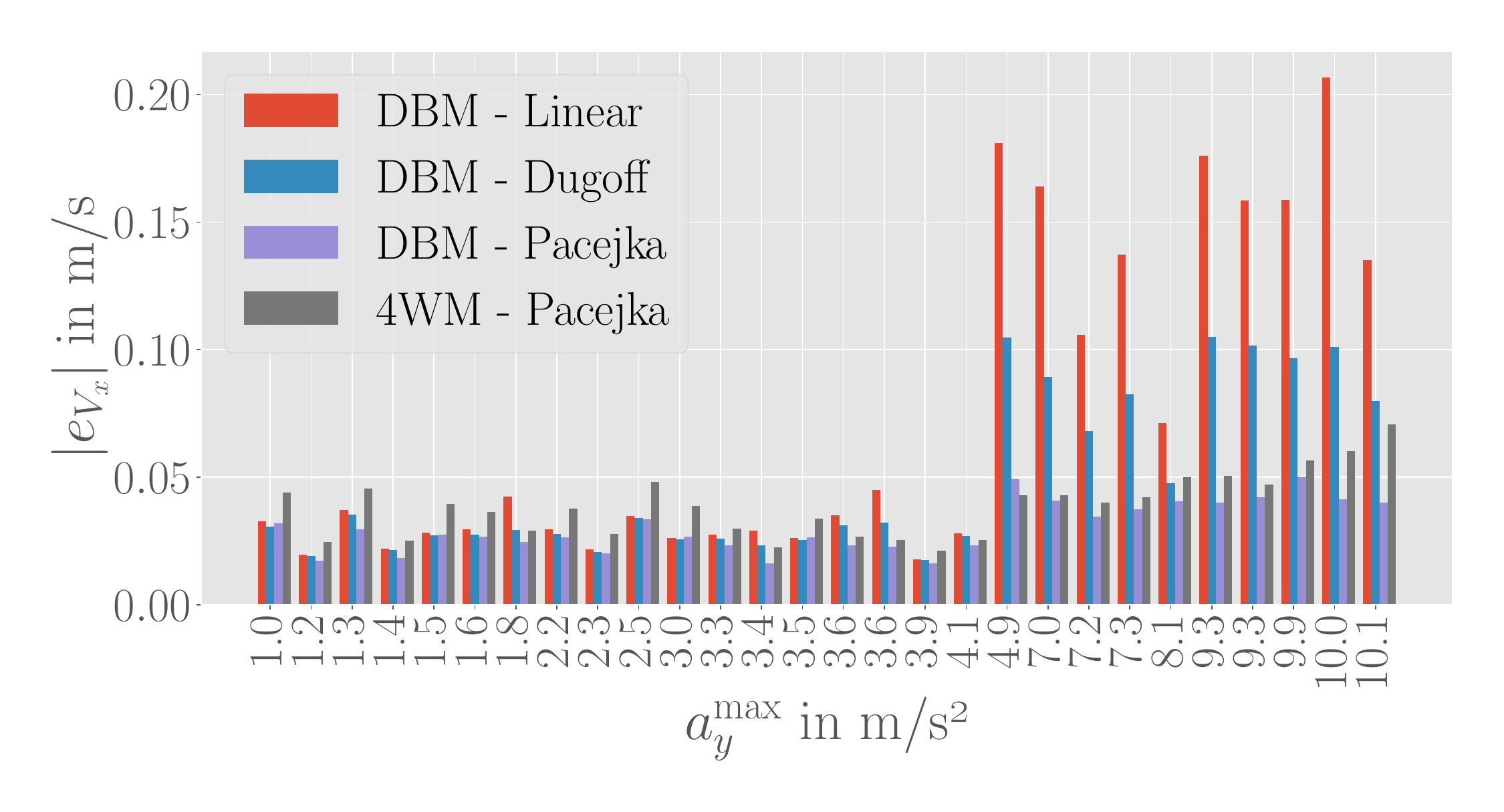}
    \caption{Mean absolute longitudinal velocity $V_x$ errors of the four models for the different trajectories. A behavior change can be seen after the $a_y^{\text{max}}=\SI{4.1}{\meter\per\second\squared}$ trajectory.}
    \label{vxComp.fig}
\end{figure}

\begin{figure}[h]
    \centering
    \includegraphics[width=\columnwidth]{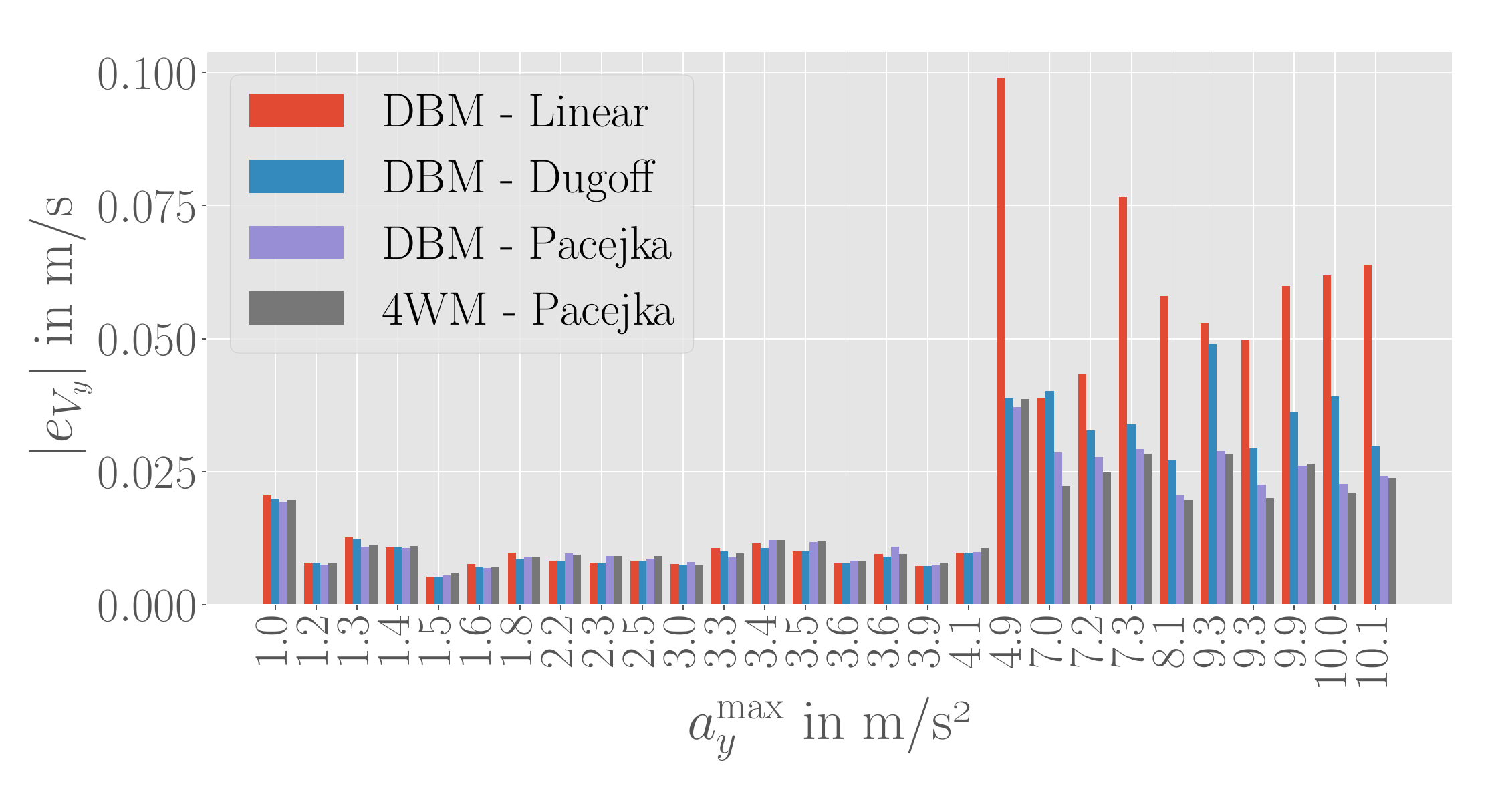}
    \caption{Mean absolute lateral velocity $V_y$ errors of the four models for the different trajectories. The same behavior change as in $V_x$ can be seen after the $a_y^{\text{max}}=\SI{4.1}{\meter\per\second\squared}$ trajectory.}
    \label{vyComp.fig}
\end{figure}

\begin{figure}[h]
    \centering
    \includegraphics[width=\columnwidth]{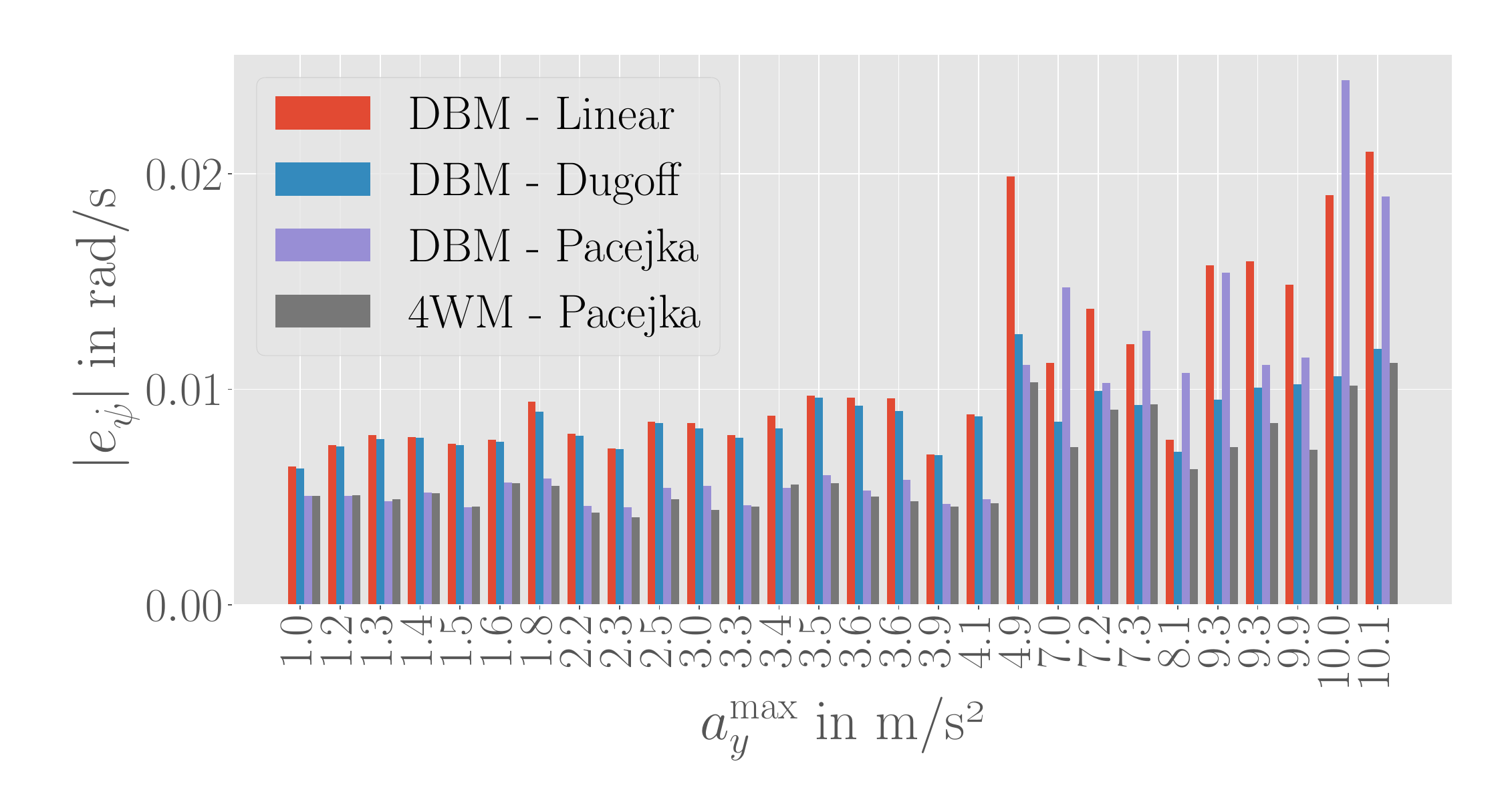}
    \caption{Mean absolute yaw rate $\dot\psi$ errors of the four models for the different trajectories. The behavior change seen before can be seen after the $a_y^{\text{max}}=\SI{4.1}{\meter\per\second\squared}$ trajectory.}
    \label{psiDotComp.fig}
\end{figure}

Having  established that a link exists between the lateral accelerations of the vehicle and the errors of the models, we aim to further analyze this link. For this purpose, the errors of all of the considered trajectories are calculated for each of the three variables. As mentioned earlier, the trajectories were picked to reflect different levels of lateral accelerations. For each of the trajectories, the mean absolute errors of each of the variables are calculated. The resulting errors for each of the trajectories are plotted with the trajectories being ordered based on their maximum lateral accelerations. The error plots for each of the variables $V_x$, $V_y$ and $\dot\psi$ are shown in Figures \ref{vxComp.fig}-\ref{psiDotComp.fig}.


\begin{figure}[h]
    \centering
    \includegraphics[width=\columnwidth]{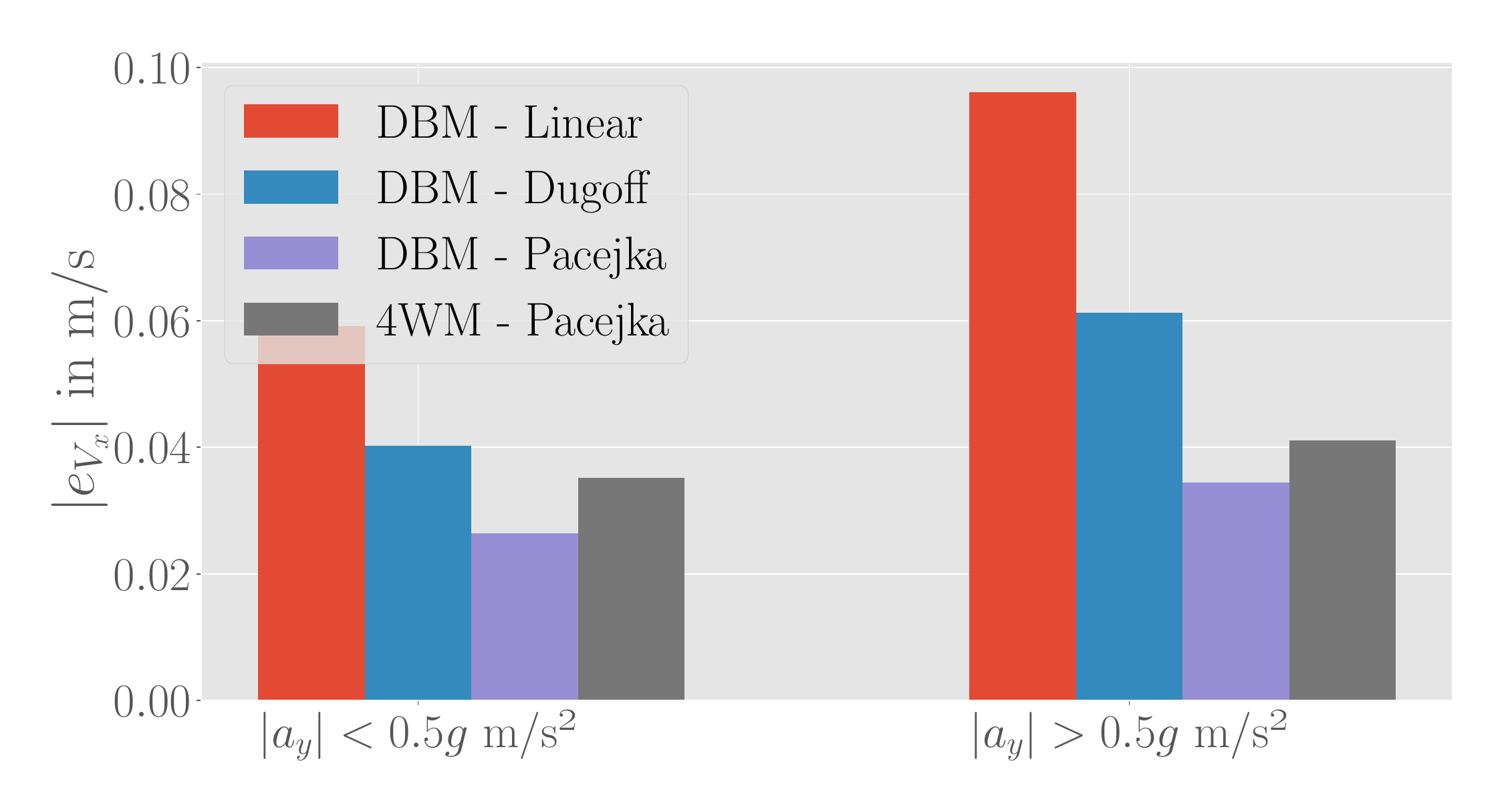}
    \caption{Mean absolute longitudinal velocity $V_x$ errors grouped to two acceleration domains. The plot shows a significant behavior change in the linear and Dugoff models between the two domains while the Pacejka models present a slight increase.}
    \label{vxCompyAy.fig}
\end{figure}

\begin{figure}
    \centering
    \includegraphics[width=\columnwidth]{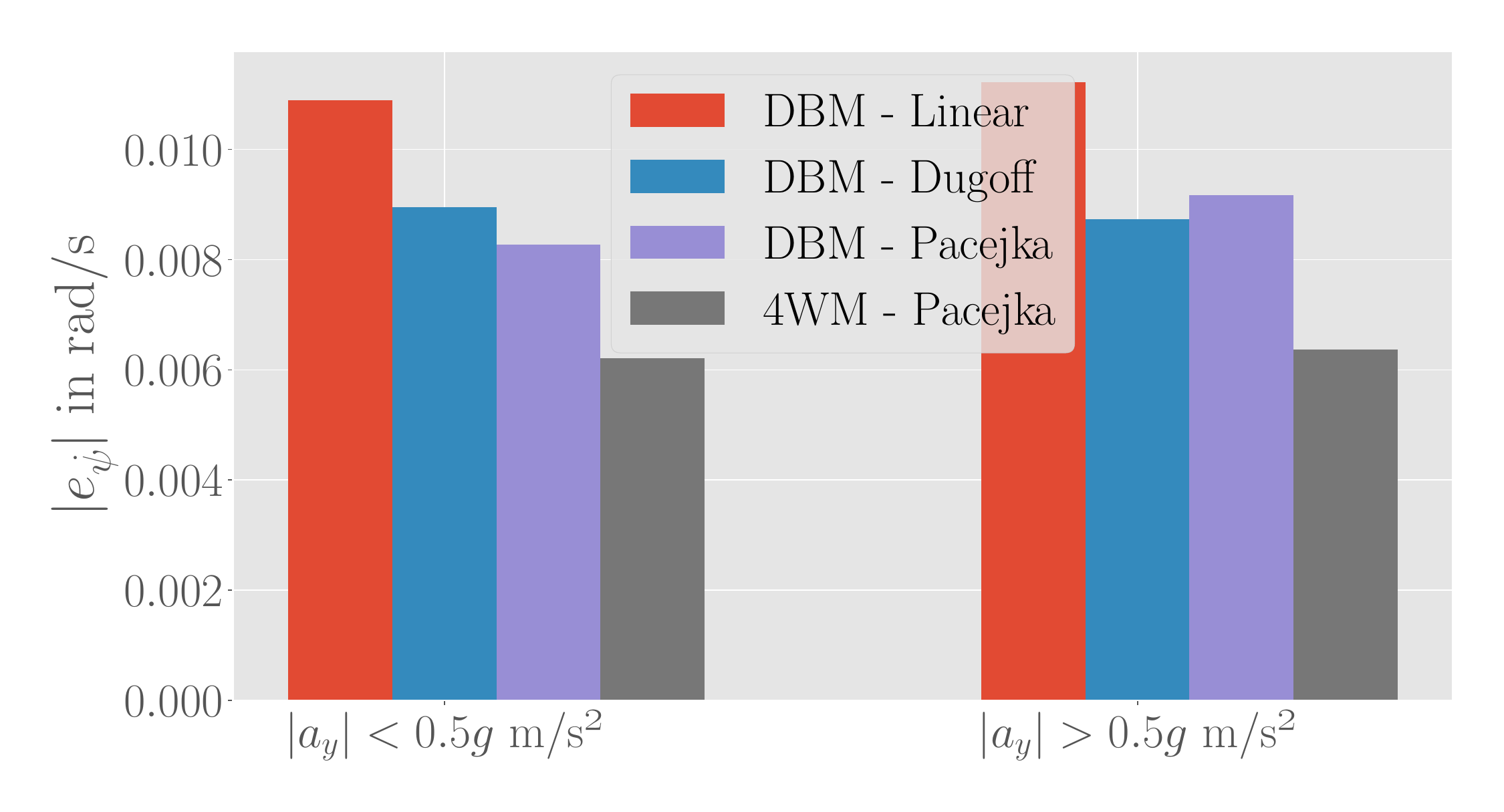}
    \caption{Mean absolute yaw rate $\dot\psi$ errors grouped to two acceleration domains. The plot shows a slight behavior change in the four models between the two acceleration domains.}
    \label{psiDotCompyAy.fig}
\end{figure}

The shown plots present a clear behavior difference for the three variables of the four models between two acceleration domains. The first acceleration domain is limited by the trajectory where $a_y^{\text{max}}=\SI{4.1}{\meter\per\second\squared}$; it refers to low acceleration maneuvers. The second acceleration domain starts at the trajectory where $a_y^{\text{max}}=\SI{4.9}{\meter\per\second\squared}$; it refers to high acceleration maneuvers. The lateral acceleration value that separates the two domains appears to be $a_y=0.5g$. The figures prove that a significant behavior change is associated with the four models whenever the lateral acceleration $a_y>0.5g$. A clearer visualization is seen when calculating the mean absolute errors for the four models for each of the two defined domains: $a_y<0.5g$ and $a_y>0.5g$: this is presented in Table \mbox{\ref{modelComp.tab}} and in Figures \mbox{\ref{vxCompyAy.fig}}-\mbox{\ref{psiDotCompyAy.fig}}. The standard deviations associated with each of the models are shown in Table \mbox{\ref{modelCompStd.tab}}.

\begin{table}[h]
\centering
\caption{Mean Absolute Errors for the three variables in the two acceleration domains. A large error increase can be seen when moving to $a_y>0.5g$. (inc.: increase). $V_x$, $V_y$ errors in $\SI{}{\meter\per\second}$; $\dot\psi$ errors in $\SI{}{\radian\per\second}$.}
{\begin{tabular}{cccccc}
\toprule
\textbf{Var.} & $a_y^{\text{max}}$ & \textbf{DBM-Lin.} & \textbf{DBM-Dug.} & \textbf{DBM-Pc.} & \textbf{4WM-Pc.}\\
\midrule
\multirow{2}{*}{$V_x$} & $<0.5g$  & 0.059 & 0.040 & 0.026 & 0.035 \\ 
\cmidrule{2-6}
& $>0.5g$ & 0.096 & 0.061 & 0.034 & 0.041 \\
\cmidrule{2-6}
\textbf{\% inc.} & & 62 & 52.5 & 23.5 & 17.1\\
\midrule
\multirow{2}{*}{$V_y$} &  $<0.5g$ & 0.020 & 0.014 & 0.013  & 0.012\\
\cmidrule{2-6}
& $>0.5g$ & 0.038 & 0.024 & 0.019 & 0.018 \\
\cmidrule{2-6}
\textbf{\% inc.} & & 90 & 71 & 31 & 50 \\
\midrule
\multirow{2}{*}{$\dot{\psi}$} & $<0.5g$ & 0.010 & 0.0088 & 0.0082 & 0.0062 \\
\cmidrule{2-6}
& $>0.5g$ & 0.011 & 0.0088 & 0.0091 & 0.0063 \\
\cmidrule{2-6}
\textbf{\% inc.} & & 10 & 11 & 11 & 16\\
\bottomrule
\end{tabular}}

\label{modelComp.tab}
\end{table}

\begin{table}[h]
\centering
\caption{Standard deviations for the errors of the three variables in the two acceleration domains. Higher standard deviations can be seen when moving to $a_y>0.5g$. $V_x$, $V_y$ in $\SI{}{\meter\per\second}$; $\dot\psi$ in $\SI{}{\radian\per\second}$.}
{\begin{tabular}{cccccc}
\toprule
\textbf{Var.} & $a_y^{\text{max}}$ & \textbf{DBM-Lin.} & \textbf{DBM-Dug.} & \textbf{DBM-Pc.} & \textbf{4WM-Pc.}\\
\midrule
\multirow{2}{*}{$V_x$} & $<0.5g$  & 0.096 & 0.044 & 0.023 & 0.039 \\ 
\cmidrule{2-6}
& $>0.5g$ & 0.112 & 0.053 & 0.034 & 0.049 \\
\midrule
\multirow{2}{*}{$V_y$} &  $<0.5g$ & 0.039 & 0.018 & 0.014  & 0.013\\
\cmidrule{2-6}
& $>0.5g$ & 0.053 & 0.022 & 0.018 & 0.016 \\
\midrule
\multirow{2}{*}{$\dot{\psi}$} & $<0.5g$ & 0.010 & 0.005 & 0.009 & 0.006 \\
\cmidrule{2-6}
& $>0.5g$ & 0.011 & 0.005 & 0.010 & 0.0068 \\
\bottomrule
\end{tabular}}

\label{modelCompStd.tab}
\end{table}

\begin{figure}[h]
    \centering
    \includegraphics[width=\columnwidth]{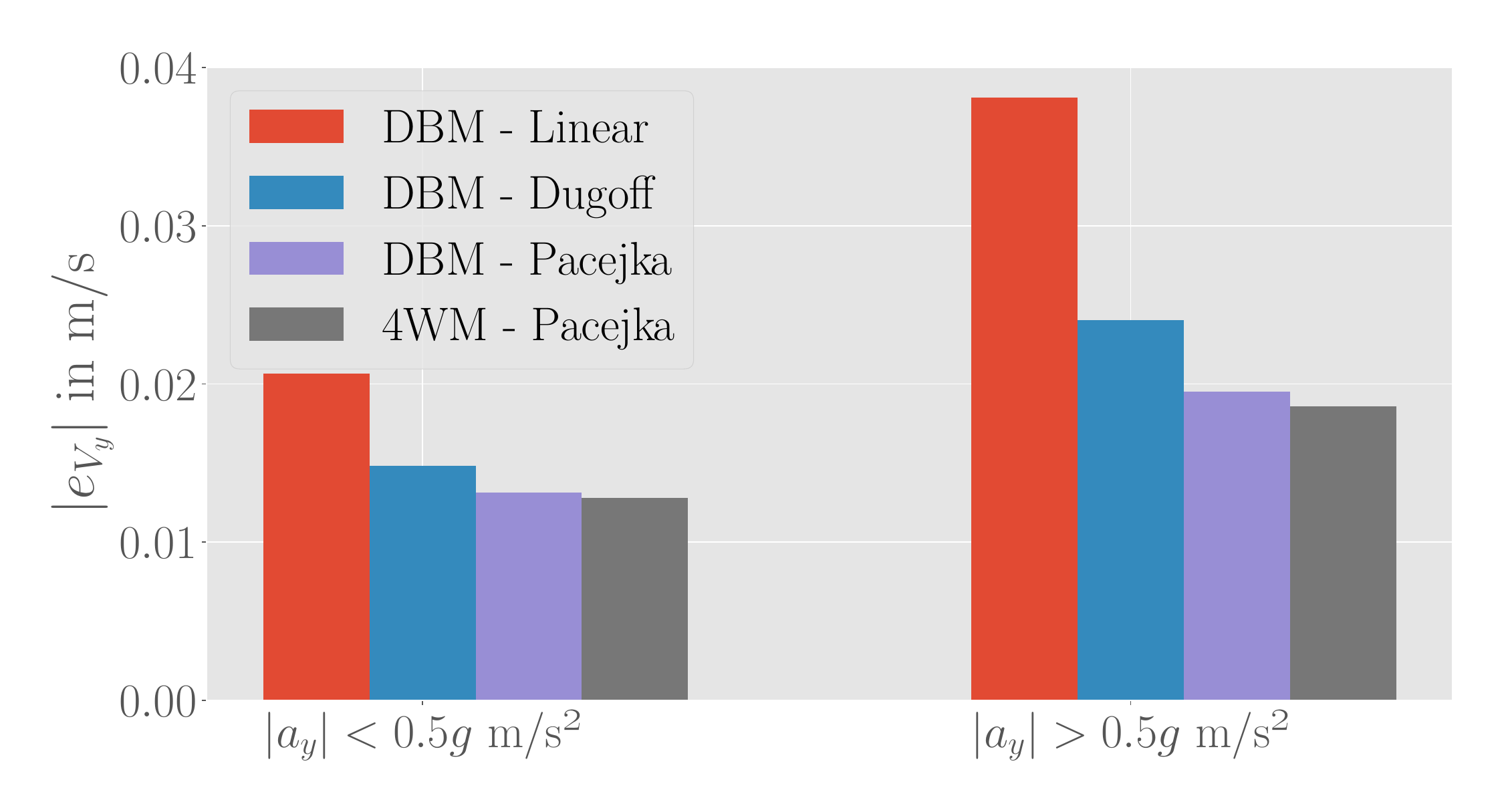}
    \caption{Mean absolute lateral velocity $V_y$ errors grouped to two acceleration domains. The plot shows a significant behavior change in the four models between the two acceleration domains.}
    \label{vyCompyAy.fig}
\end{figure}

The presented results show a large difference in performance between the two acceleration domains, especially for the $V_x$ and $V_y$ variables. For the longitudinal velocity $V_x$, the linear-based and Dugoff-based models show a significant error increase between the two presented domains while the Pacejka-based models show more robustness to higher accelerations. For the lateral velocity $V_y$, all of the four models show a significant error increase between the two domains. While for the yaw rate $\dot\psi$, the Dugoff model presents the same behavior between both domains while the other models present an increase in the model errors. The effect of the lateral acceleration domain on the performance of each model can be seen as well in the standard deviations variations: higher lateral accelerations are associated with higher deviations.

The presented analysis highlights the role of the tire model used in the model's accuracy, having both Pacejka-based models delivering close performance despite the vehicle model change.

It is clear that the performance of the models differs between the two acceleration domains. Moving to the higher acceleration domain is associated with higher errors: this is due to the inability of the models to accurately represent the highly nonlinear vehicle behavior, especially the linear model.  

\subsubsection{Which model is valid?}
Having compared the different models and deduced their performance change in the different acceleration domains, a question that arises is the validity of each model. The presented results show that the four models are able to provide low, close errors for the low acceleration domain; the problem lies in the high acceleration domain. Beyond $a_y=0.5g$, the linear-based model shows a significant error increase and presents almost 2 times higher errors than the Pacejka-based models. The Dugoff-based model shows an increase in errors, but keeps a closer performance to the Pacejka-based models. Both Pacejka-based models show the most robustness to acceleration changes and are able to deliver the lowest errors in all cases. The linear-based model seems to lose validity in the higher acceleration domain, while the Pacejka-based models seem the most valid; the Dugoff-based model is between both performances.

In the presented data, the relationship between the model validity and the acceleration domain of the vehicle is clear. Operating any of the models in the high acceleration domain results in higher errors. Though, the performance decline is not the same between the models. 

When employing either of these models in a real vehicle control application, the needed accuracy must be considered: this will determine if the model is valid in the needed context. To further illustrate the effects of model errors and validity on control-related applications, we implement the presented models in state estimators next. 

\section{Real Vehicle Observing}\label{realObserving.sec}
Having identified the modeling errors associated with the presented vehicle and tire models, and the link between these errors and the lateral acceleration domain of the vehicle, we aim to prove the effects of using each of the models in a state observing application for different lateral acceleration considerations. The Extended Kalman Filter (EKF) is then implemented for our state observing application. The EKF is a nonlinear extension to the Kalman Filter, and is well known in the literature for vehicle observing applications (e.g. \mbox{\cite{chen_sideslip_2008, hrgetic_vehicle_2014, kim_vehicle_2018, zong_dual_2013})}. The EKF will be used to estimate the longitudinal and lateral velocities ($V_x$, $V_y$) and the yaw rate ($\dot\psi$) of the vehicle.

In what follows, the EKF algorithm is defined in Section \ref{EKF.ssec}, its application to the above-presented models is detailed in Section \ref{EKFVehicles.ssec} and its estimation results are presented in Section \ref{EKF_res.ssec}.

After analyzing the results of the four model-based observers, a brief comparison with a learned observer is presented in Section \ref{EKF_learned.ssec}.

\subsection{The EKF Algorithm}\label{EKF.ssec}
The EKF is a state observing algorithm that takes into consideration the process and measurement noise representing the model errors and mismatches and the sensor measurement errors respectively. It involves a prediction step where the current state and covariances of the system are calculated based on the previously estimated state and the current inputs and covariances and an update step where the current measurements are used to correct the calculated state and covariances. 

Consider the following discrete nonlinear system subject to both process noise and measurement noise:

\begin{subequations}
\begin{eqnarray}
{    Z_k = f(Z_{k-1},U_k) + w_k} \\
{m_k = g(Z_k) + v_k}
\end{eqnarray}
\end{subequations}
$Z_k$ being the state of the system, $U_k$ being the inputs to the system, $m_k$ being the measurement. Noises are assumed to be Gaussian: $w_k$ being the process noise with covariance $Q_k$ and $v_k$ being the measurement noise with covariance $R_k$.\\
The equations of the EKF that uses the presented system are the following:
\begin{subequations}
\begin{eqnarray}
&\text{Predict}& \nonumber\\
\hat{Z}_{k|k-1}&=&f(\hat{Z}_{k-1|k-1}, U_k)\\
{P_{k|k-1}}&=&F_k\hat{P}_{k-1|k-1}F_k^T+Q_k\\
&\text{Update}& \nonumber\\*
\Tilde{m_k} &=& m_k-g(\hat{Z}_{k|k-1})\\*
S_k &=& G_kP_{k|k-1}G_k^T+R_k\\*
K_k &=& P_{k|k-1}G_k^TS_k^{-1} \\*
\hat{Z}_{k|k} &=& \hat{Z}_{k|k-1} + K_k\Tilde{m}_k\\*
P_{k|k} &=& (I-K_kG_k)P_{k|k-1}
\end{eqnarray}
\end{subequations}
$F_k$ is the state transition matrix and $G_k$ is the observation matrix defined as:
\begin{subequations}
\begin{eqnarray}
{ F_k=\frac{\partial f}{\partial Z}\bigg\rvert_{\hat{Z}_{k-1|k-1}, U_k} }\\{
     G_k = \frac{\partial g}{\partial Z}\bigg\rvert_{\hat{Z}_{k-1|k-1}}}
\end{eqnarray}
\end{subequations}
The EKF is applied to the vehicle models in the next section. 

\subsection{Application to Vehicle Models}\label{EKFVehicles.ssec}
Having defined the EKF algorithm, it is employed with the presented models in Section \ref{soa.sec} for vehicle state estimation. The state $Z$ to be observed consists of the longitudinal and lateral velocities and the yaw rate: $Z=\begin{bmatrix}
    V_x & V_y & \dot\psi
\end{bmatrix}$. The used measurements $m$ are the longitudinal and lateral accelerations and yaw rate provided by the standard car sensors: $m=\begin{bmatrix}
    \Tilde{a_x} & \Tilde{a_y} & \Tilde{\dot\psi}
\end{bmatrix}$. 
The four employed EKFs are based on the four-wheel model with the Pacejka tire model, the dynamic bicycle model with the linear tire model, the dynamic bicycle model with the Dugoff tire model, and the dynamic bicycle model with the Pacejka tire model. 

For the four-wheel vehicle model, the observer implemented is inspired by the work done in \cite{katriniok_adaptive_2016}. The considered implementation neglects the road slope and bank angles, and the roll and pitch dynamics.  The state update equations are then defined in Equations (\ref{4wm1.eq}), (\ref{4wm2.eq}), (\ref{4wm3.eq}). The vertical forces are defined in Equations (\ref{Fz.eq}). The measurement functions are derived from the state update equations using Equations (\ref{acc.eq}) and are defined as:
\begin{subequations}
    \begin{eqnarray}
        \Tilde{a_x} &=& \frac{1}{M_T} ((F_{xp}^{fl}+F_{xp}^{fr})\cos\delta- \\ &&(F_{yp}^{fl}+F_{yp}^{fr})\sin\delta +  F_{xp}^{rl}+F_{xp}^{rr}) \nonumber\\
        \Tilde{a_y} &=& \frac{1}{M_T} ((F_{yp}^{fl}+F_{yp}^{fr})\cos\delta + \\ && (F_{xp}^{fl}+F_{xp}^{fr})\sin\delta +F_{yp}^{rl}+F_{yp}^{rr}) \nonumber\\
        \Tilde{\dot\psi} &=& \dot\psi
        \end{eqnarray}
\end{subequations}

For the dynamic bicycle model, the state update equations are defined by Equations (\ref{DBM.eq}). The vertical forces used for the Dugoff tire model are as well defined using Equations (\ref{Fz.eq}). The measurement functions are defined as:
\begin{subequations}
    \begin{eqnarray}
        \Tilde{a_x} &=& \frac{1}{M_T} (F_{xp}^{f}\cos\delta -F_{yp}^{f}\sin\delta + F_{xp}^{r})\\
        \Tilde{a_y} &=& \frac{1}{M_T} (F_{yp}^{f}\cos\delta + F_{xp}^{f}\sin\delta + F_{yp}^{r})\\
        \Tilde{\dot\psi} &=& \dot\psi
        \end{eqnarray}
\end{subequations}

In each of the observers, the process and measurement covariances are changed for each trajectory. Knowing that the process covariance is based on process errors and that the ground truth sensor provides the data for the state variables, the process covariance matrix in this work is calculated for each trajectory based on the error calculation present in Algorithm \mbox{\ref{1.alg}}. In the same way, the measurement errors are calculated using the same algorithm considering the difference between the SARA vehicle sensors and the ground truth sensor for each trajectory. The choice of calculating the errors for each scenario beforehand and applying corresponding covariances gives an advantage to the used observer as it will be adapted to the errors of each specific scenario; this will highlight the observer errors with disregard to the covariances tuning, allowing to interpret the model validity impact on the observer's performance.

The defined observers are tested on the same trajectories defined above and their estimations are analyzed next.

\subsection{Results and Analysis}\label{EKF_res.ssec}
Having presented the EKF observer and its application using the different vehicle models, the four observers are compared in this section. As we seek to analyze the effects of model validity on the state observation of the vehicle, we follow an analysis procedure similar to the one presented in the previous section. The different mean absolute errors will be presented for each of the trajectories, knowing that in this case the errors are defined as: 
\begin{equation}
    e_k = |Z_k^{\text{estimated}} - Z_k^{\text{gt}}|
\end{equation}
As opposed to the previous error calculation algorithm, in this case, the observer algorithm is run for the whole trajectory with its initial state only being fed from the ground truth values. 

\begin{figure}[h]
    \centering
    \includegraphics[width=\columnwidth]{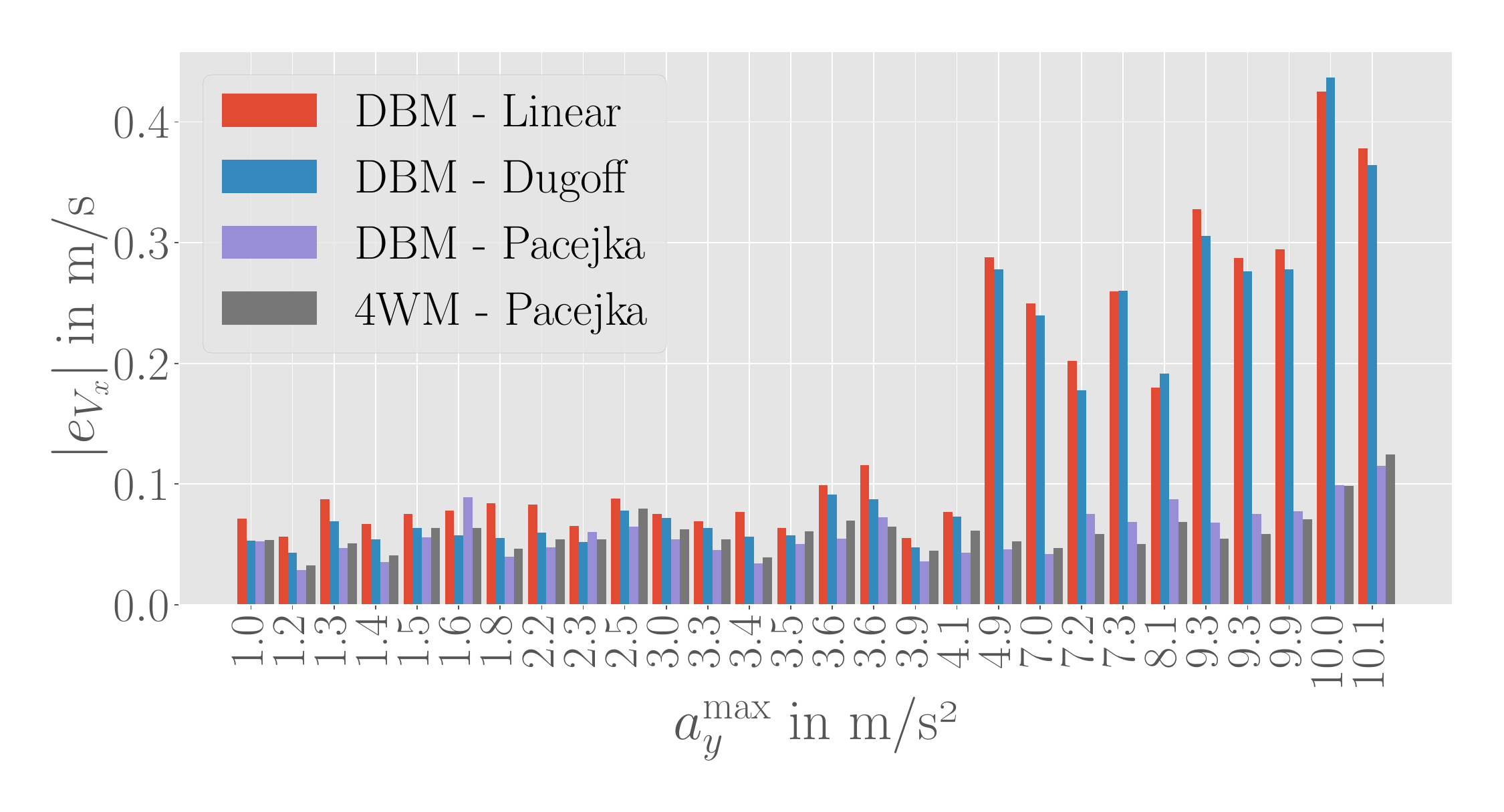}
    \caption{Mean absolute errors of the estimations of the longitudinal velocities $V_x$ by the four EKFs. The plot shows that a behavior difference is present for the different models between $a_y^{\text{max}}<0.5g$ and $a_y^{\text{max}}>0.5g$ with more robustness to the Pacejka-based models.}
    \label{vx_obs.fig}
\end{figure}

\begin{figure}[h]
    \centering
    \includegraphics[width=\columnwidth]{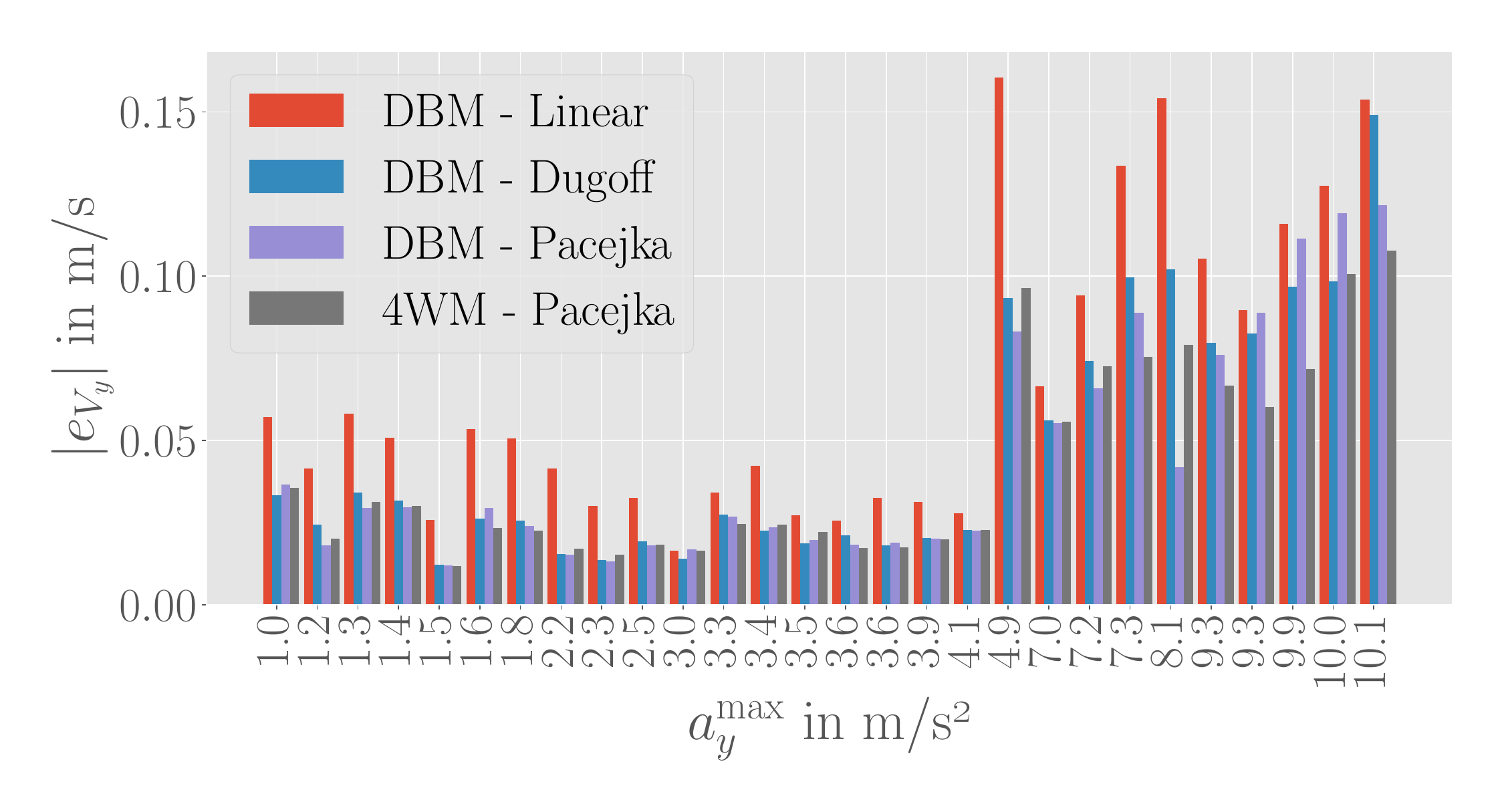}
    \caption{Mean absolute errors of the estimations of the lateral velocities $V_y$ by the four EKFs. The plot shows that a behavior difference is present for the four observers between $a_y^{\text{max}}<0.5g$ and $a_y^{\text{max}}>0.5g$.}
    \label{vy_obs.fig}
\end{figure}

\begin{figure}[h]
    \centering
    \includegraphics[width=\columnwidth]{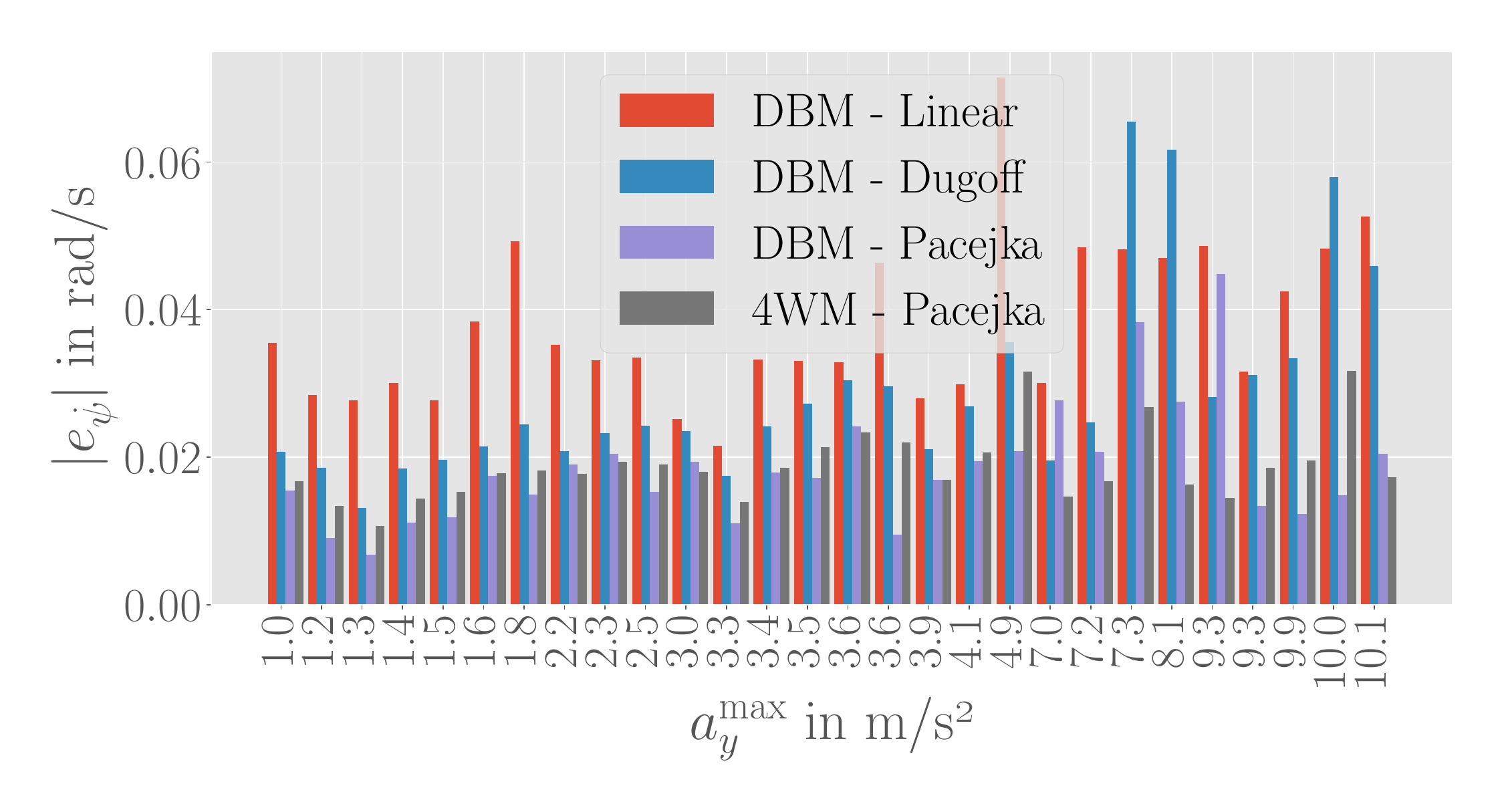}
    \caption{Mean absolute errors of the estimations of the yaw rate $\dot\psi$ by the four EKFs. The plot shows that a small behavior difference is present for the four observers between $a_y^{\text{max}}<0.5g$ and $a_y^{\text{max}}>0.5g$.}
    \label{psidot_obs.fig}
\end{figure}

After running the observers for each of the trajectories, their mean absolute errors are calculated and plotted for each variable with the trajectories being sorted in an increasing lateral acceleration order as done before. Figures \ref{vx_obs.fig}-\ref{psidot_obs.fig} show the error plots of the longitudinal velocity estimations, lateral velocity estimations, and the yaw rate estimations respectively. It should be noted that the magnitude of the error values should not be compared with that of the plots of the previous section due to the difference in the error computation procedure. 


The shown plots prove that the difference in behavior between acceleration domains proven in the previous section has direct effects on the observing accuracy for all of the four models. Though, for the yaw rate, the effects are minimal: this is due to the adaptation of the observers to the provided yaw rate measures. It is remarkable that the Pacejka-based observers show no significant error increase (error decrease can be seen in some cases) for the longitudinal velocity estimation: this proves the ability of the observers to take advantage of the robustness of the model to deliver accurate estimations; this is not the case for the other observers. 

\begin{figure}[h]
    \centering
    \includegraphics[width=\columnwidth]{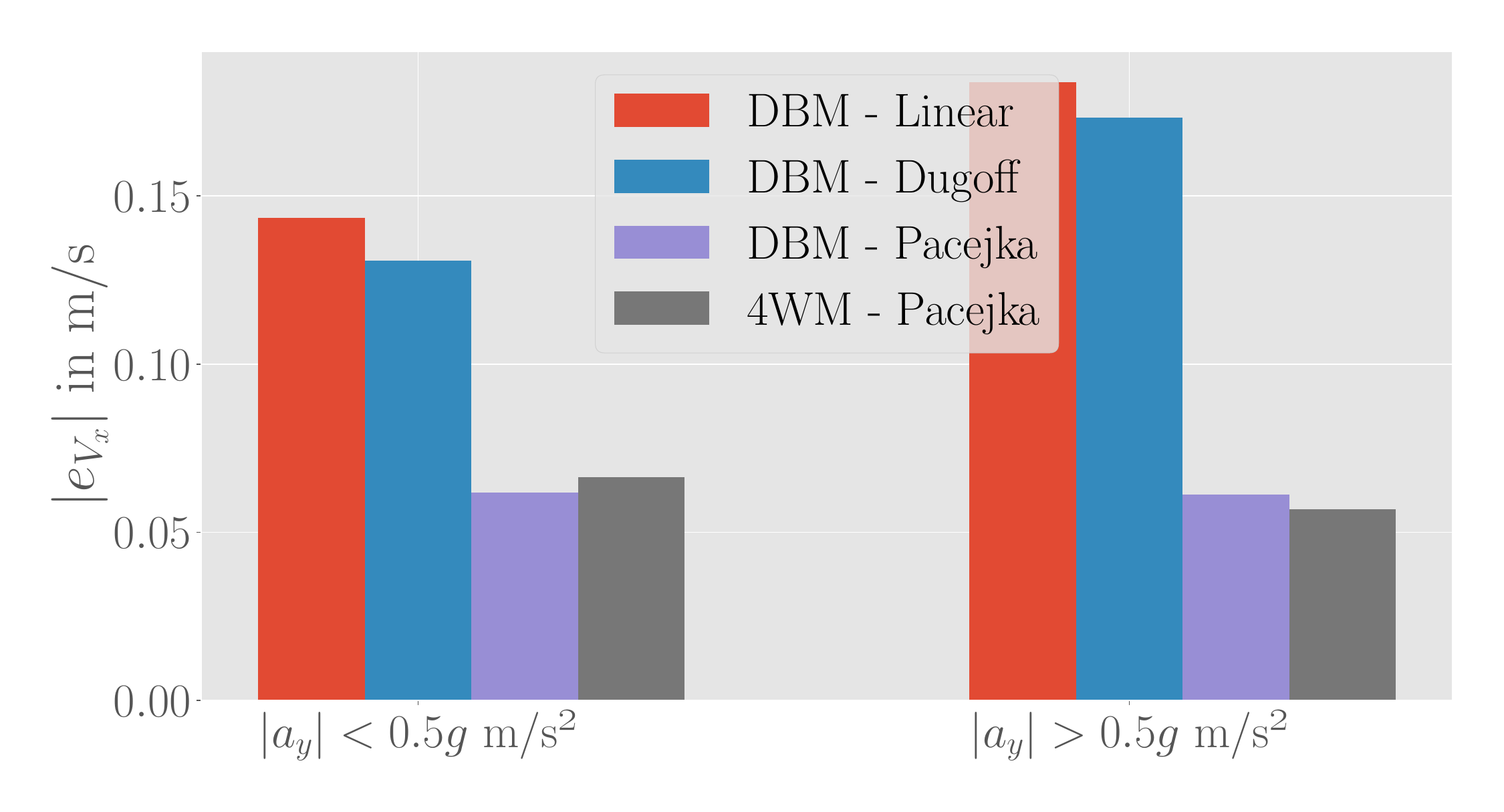}
    \caption{Comparison between the mean absolute longitudinal velocity errors of the observers for the two acceleration domains. The linear-based and Dugoff-based observers show error increase between the two domains while the Pacejka-based observers are able to deliver low errors in both domains.}
    \label{obsCompVx.fig}
\end{figure}

\begin{figure}[h]
    \centering
    \includegraphics[width=\columnwidth]{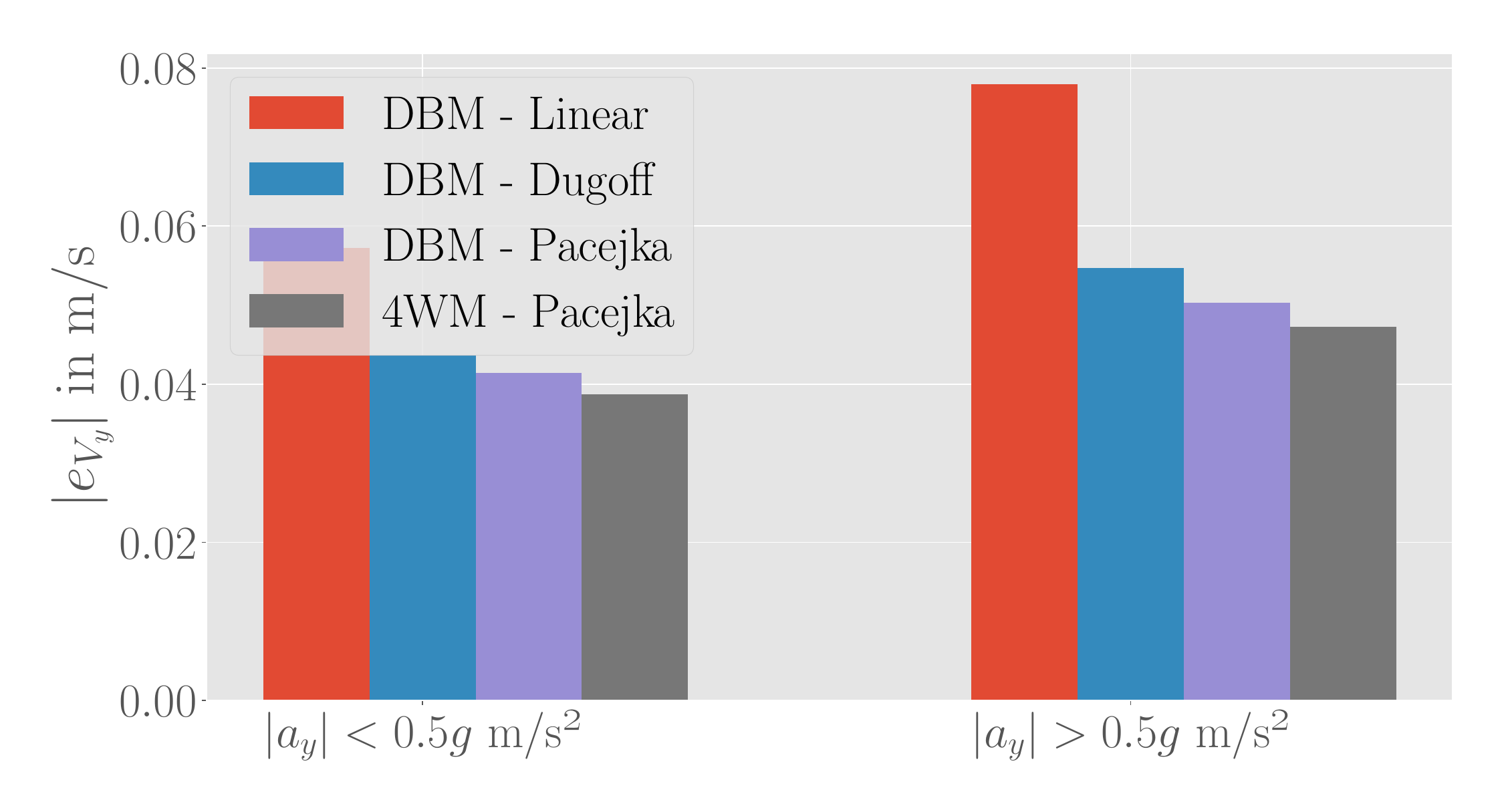}
    \caption{Comparison between the mean absolute lateral velocity errors of the observers for the two acceleration domains. The four observers show error increase between the two domains while the four-wheel Pacejka-based observer shows the lowest errors.}
    \label{obsCompVy.fig}
\end{figure}

\begin{figure}[h]
    \centering
    \includegraphics[width=\columnwidth]{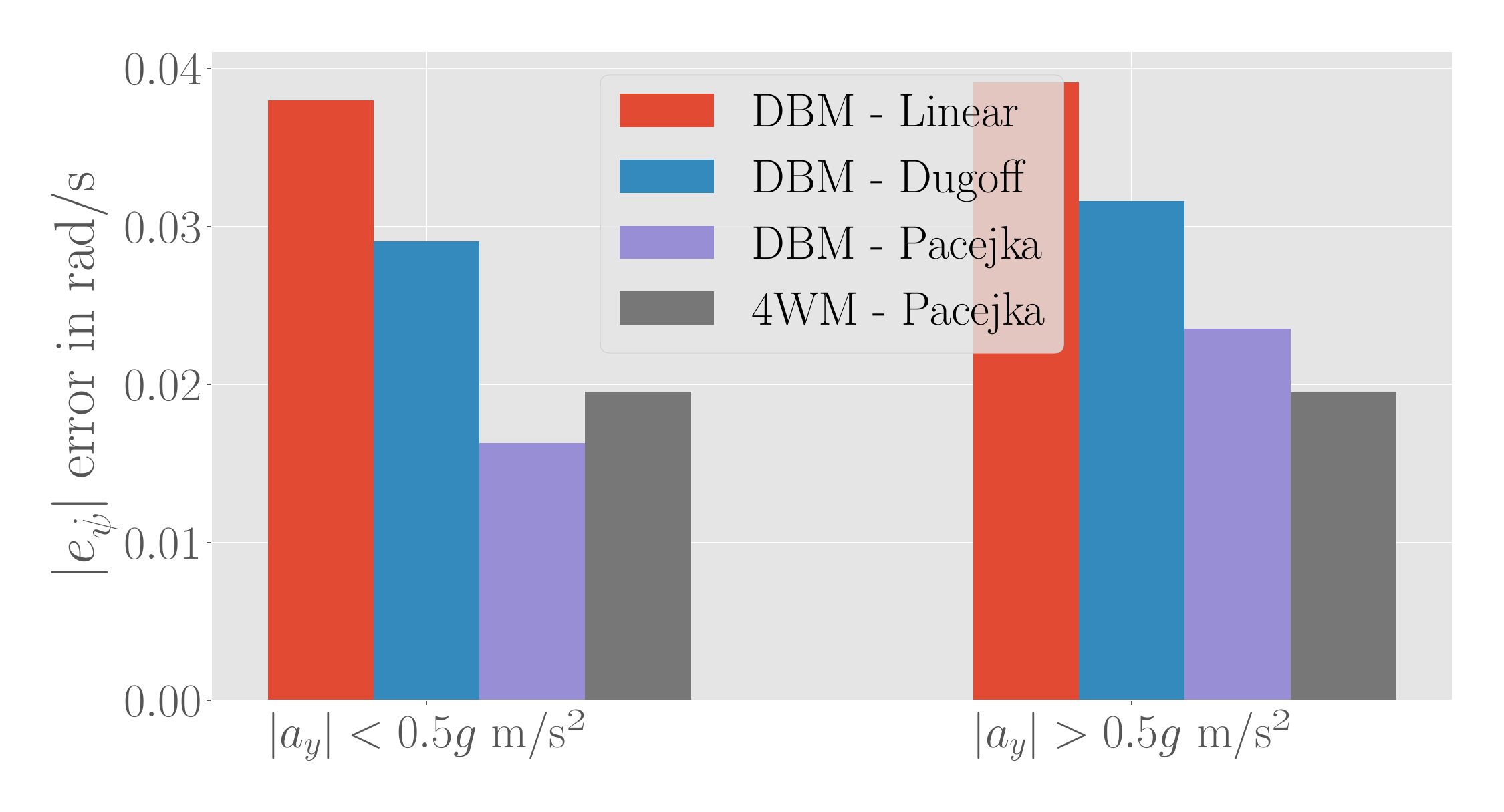}
    \caption{Comparison between the mean absolute yaw rate errors of the observers for the two acceleration domains. The linear-based, Dugoff-based and the four-wheel Pacejka observers show a slight change in behavior between the two domains: this is due to the observer taking advantage of the yaw rate measurements, this is not the case for the Pacejka-based bicycle observer. Both Pacejka-based observers show the lowest errors.}
    \label{obsCompPsidot.fig}
\end{figure}

To further investigate the performance difference of the four observers for the two acceleration domains, the mean absolute errors for the combined estimations for each of the two acceleration domains are calculated: this is shown in Table \mbox{\ref{obsComp.tab}} and Figures \mbox{\ref{obsCompVx.fig}-\ref{obsCompPsidot.fig}}. The shown errors prove the relationship between the model validity and the performance of the observers. The linear and Dugoff based observers show a clear increase in estimation errors for longitudinal and lateral velocities beyond the $a_y=0.5g$ acceleration point. The Pacejka-based observers take advantage of the model accuracy, already shown in the previous section, and are able to deliver low errors in both domains for the longitudinal velocity; their errors are the lowest for the longitudinal, lateral velocity and yaw rate estimations.

\begin{table}[h]

\centering
\caption{Mean Absolute Errors for the three variables estimations in the two acceleration domains. An error increase can be seen when moving to $a_y>0.5g$ except for the Pacejka-based observer in $V_x$ estimations. (inc.: increase). $V_x$, $V_y$ errors in $\SI{}{\meter\per\second}$; $\dot\psi$ errors in $\SI{}{\radian\per\second}$.}
\begin{tabular}{cccccc}
\toprule
\textbf{Var.} & $a_y^{\text{max}}$ & \textbf{DBM-Lin.} & \textbf{DBM-Dug.} & \textbf{DBM-Pc.} & \textbf{4WM-Pc.}\\
\midrule
\multirow{2}{*}{$V_x$} & $<0.5g$  & 0.14 & 0.13 & 0.061 & 0.066 \\ 
\cmidrule{2-6}
& $>0.5g$ & 0.18 & 0.17 & 0.061 & 0.056 \\
\cmidrule{2-6}
\textbf{\% inc.} & & 28 & 30 & 1 & -15\\
\midrule
\multirow{2}{*}{$V_y$} &  $<0.5g$ & 0.057 & 0.043 & 0.041  & 0.038\\
\cmidrule{2-6}
& $>0.5g$ & 0.077 & 0.054 & 0.050 & 0.047 \\
\cmidrule{2-6}
\textbf{\% inc.} & & 35 & 25 & 21 & 23 \\
\midrule
\multirow{2}{*}{$\dot{\psi}$} & $<0.5g$ & 0.037 & 0.029 & 0.016 & 0.019 \\
\cmidrule{2-6}
& $>0.5g$ & 0.039 & 0.031 & 0.023 & 0.019 \\
\cmidrule{2-6}
\textbf{\% inc.} & & 5.4 & 6.8 & 43 & 2.6\\
\bottomrule
\end{tabular}

\label{obsComp.tab}
\end{table}




The presented estimation errors prove the effect of model validity in state observing applications. The four models show a change of behavior between the two acceleration domains shown in the previous sections. Although the EKF is able to take advantage of both the model and measurement properties as shown in the Pacejka estimations for $V_x$ or in the yaw rate estimations, the fact that it is model-based makes it reliant on the properties of the used model, thus its validity domain. Using a valid model in observing applications is important for accurate state knowledge. 

After showing the effects of model validity on state observing, the question that arises is the ability of learned observers to overcome the disadvantages of model-based ones, and to learn the dynamics of the vehicle in different scenarios, ensuring a larger validity domain. To answer this question a learning-based observing solution is tested next. 

\subsection{Comparison with a Learned Observer}\label{EKF_learned.ssec}
Having presented model validity issues and their effects on the accuracy of state observers, we investigate the accuracy of learned observers in this section. Learned observers have gained popularity in recent years for their ability to provide accurate estimations in a wide range of maneuvers. They are based on a neural network architecture that learns the dynamics of the model from the available measurement inputs and delivers accurate estimation outputs. The learned observers encountered in the literature are of two types: hybrid or fully learned observers. In the first case, the used observer is a combination of model-based techniques and neural networks, as the KalmanNet \cite{revach_kalmannet_2021,escoriza_data-driven_2021} that consists of a Kalman filter in which the Kalman gain calculation is replaced by recurrent neural networks; another example is combining a sliding-mode observer with neural networks as in \cite{song_vehicle_2021}. In the second case, a set of neural networks is used, in various architectures, to perform estimations as done in \cite{srinivasan_end--end_2020, zhang_reliable_2021, ghosn_robust_2023}. 

\begin{figure}[h]
    \centering
    \includegraphics[width=\columnwidth]{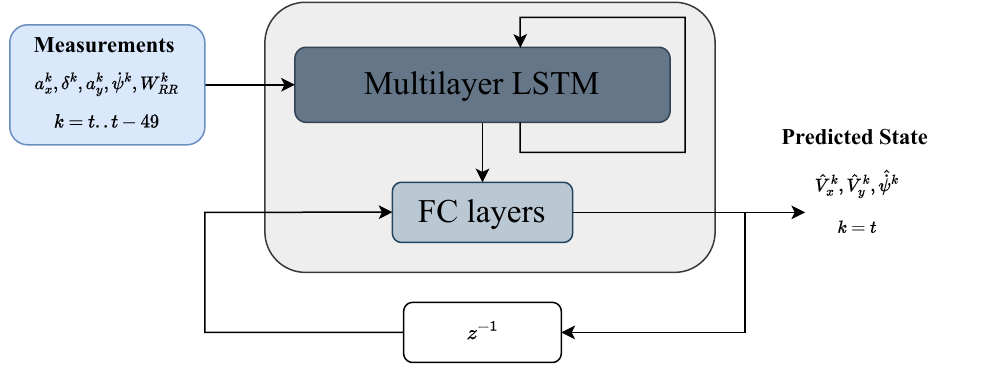}
    \caption{Learned Observer Architecture.}
    \label{arch.fig}
\end{figure}

To compare the performance of a learned observer to the model-based ones introduced before, the work done in \cite{ghosn_robust_2023} is implemented and applied on the data collected from the vehicle. The selected work was shown to outperform different model-based and learning-based observers including the KalmanNet. The architecture of the network is shown in Fig. \ref{arch.fig}. It makes use of the 50 previous measurements containing the longitudinal and lateral accelerations, the steering angle, the wheel speeds and the yaw rate and delivers the longitudinal and lateral velocities and yaw rate estimations. It consists of Long Short-Term Memory (LSTM) cells.

\begin{figure}[h]
    \centering
    \includegraphics[width=\columnwidth]{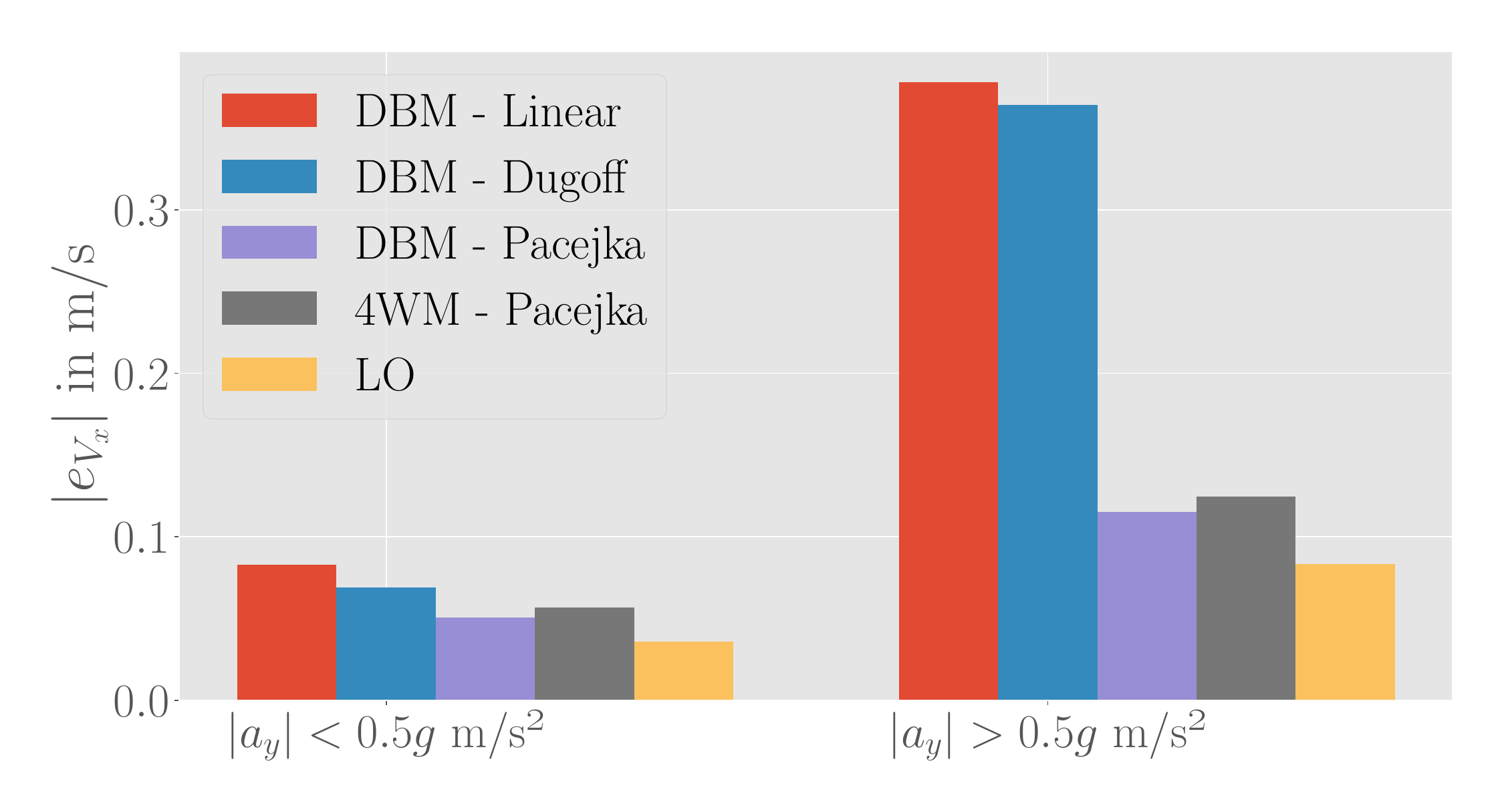}
    \caption{Mean absolute longitudinal velocity errors of the model-based and the learned observer for the two acceleration domains. A performance difference can be seen for all of the observers but the learned observer is able to deliver the lowest errors.}
    \label{LO_Vx.fig}
\end{figure}

\begin{figure}[h]
    \centering
    \includegraphics[width=\columnwidth]{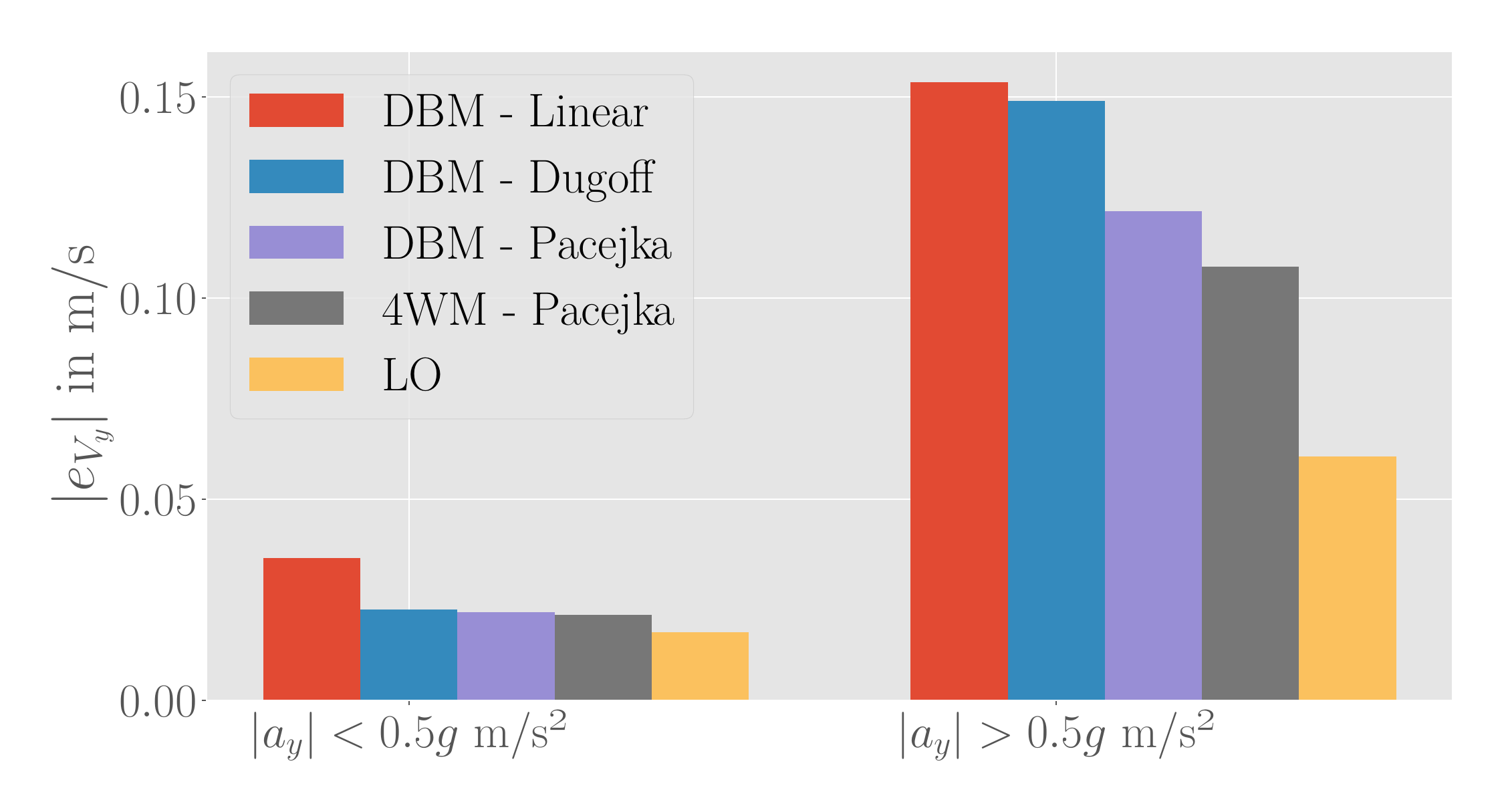}
    \caption{Mean absolute lateral velocity errors of the model-based and the learned observer for the two acceleration domains. A performance difference can be seen for all of the observers but the learned observer is able to deliver the lowest errors.}
    \label{LO_Vy.fig}
\end{figure}

\begin{figure}[h]
    \centering
    \includegraphics[width=\columnwidth]{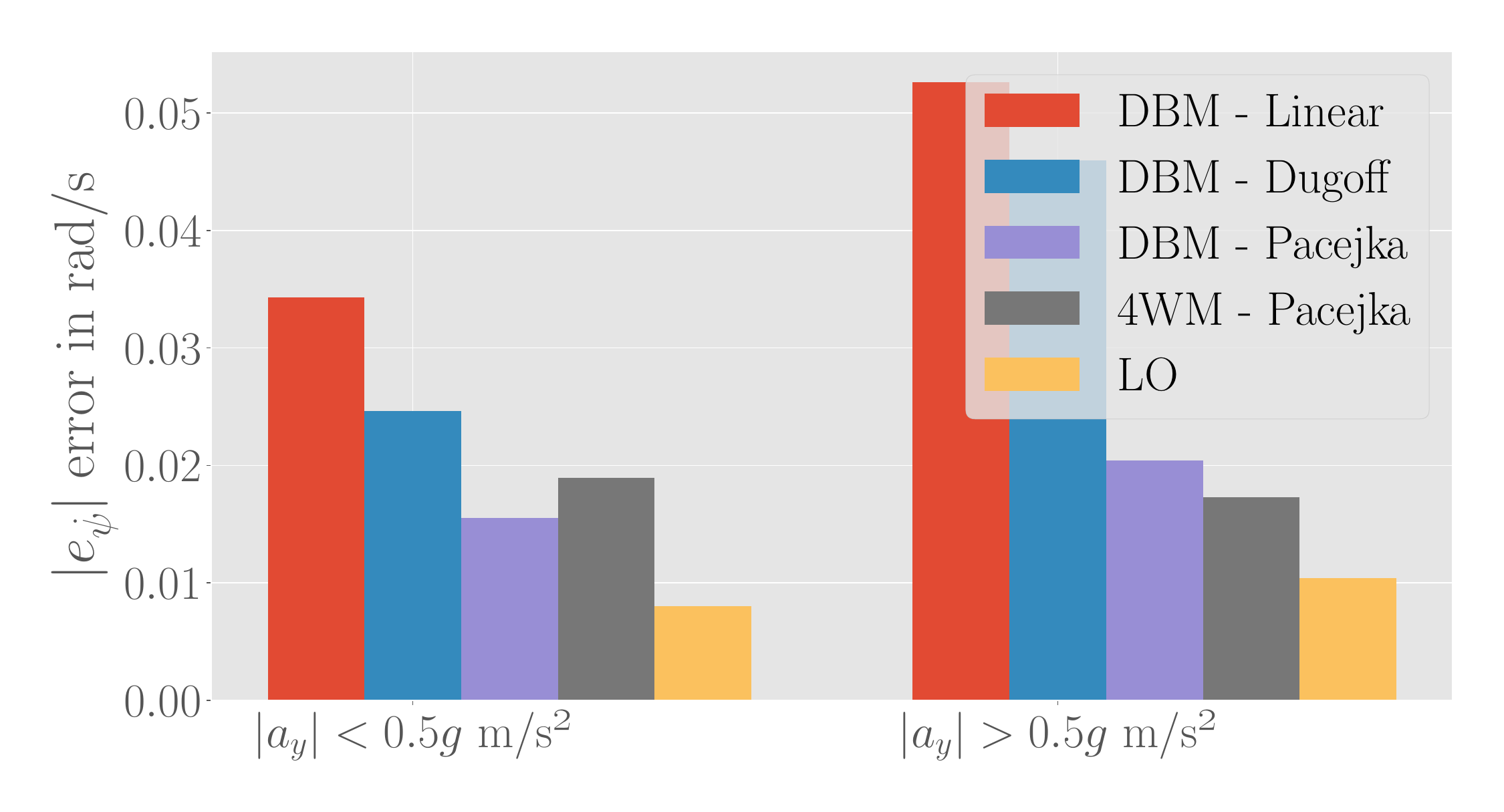}
    \caption{Mean absolute yaw rate errors of the model-based and the learned observer for the two acceleration domains. A performance difference can be seen for all of the observers but the learned observer is able to deliver the lowest errors.}
    \label{LO_Pd.fig}
\end{figure}

To be able to train the network, some of the trajectories used in the model testing presented before are added to the training set. The collected data set presented in Section \ref{realData.sec} is subject to a 60\%-20\%-20\% train-validation-test split.  After training the learned observer (LO), it is applied to the test data and its mean absolute errors are split, as done before, to two acceleration domains, along with the errors of the model-based observers: this is presented in Table \ref{LOComp.tab} and Figures \ref{LO_Vx.fig}-\ref{LO_Pd.fig}.   

\begin{table}[h]
\centering
\caption{Mean Absolute Errors for the three variables estimations in the two acceleration domains. A comparison with the learned observer (LO) is presented with a percentage difference factor (\% diff.). The table shows a close performance with the Pacejka-based observer and a bigger difference with the other observers. $V_x$, $V_y$ errors in $\SI{}{\meter\per\second}$; $\dot\psi$ errors in $\SI{}{\radian\per\second}$.}
\begin{tabular}{ccccccc}
\toprule
\textbf{Vr.} & $a_y^{\text{max}}$ & \textbf{DBM-L.} & \textbf{DBM-D.} & \textbf{DBM-P.} & \textbf{4WM-P.} & \textbf{LO}\\
\midrule
\multirow{4}{*}{$V_x$} & $<0.5g$  & 0.08 & 0.068 & 0.050 & 0.056 & 0.036\\ 
& \textbf{\% diff.} & 122 & 88 & 38 & 55 & -\\
\cmidrule{2-7}
& $>0.5g$ & 0.37 & 0.36 & 0.11 & 0.12 & 0.08\\
& \textbf{\% diff.} & 362 & 350 & 37 & 50 & -\\
\midrule
\multirow{4}{*}{$V_y$} &  $<0.5g$ & 0.035 & 0.022  & 0.022 & 0.021 & 0.016\\
& \textbf{\% diff.} & 118 & 37 & 36 & 31 & - \\
\cmidrule{2-7}
& $>0.5g$ & 0.15 & 0.14 & 0.12 & 0.10 & 0.06 \\
& \textbf{\% diff.} & 150 & 133 & 100 & 66 & - \\
\midrule
\multirow{4}{*}{$\dot{\psi}$} & $<0.5g$ & 0.034 & 0.024 & 0.015 & 0.018 & 0.008\\
& \textbf{\% diff.} & 325 & 200 & 87 &  125 & -\\
\cmidrule{2-7}
& $>0.5g$ & 0.052 & 0.045 & 0.020 & 0.0172 & 0.011 \\
    & \textbf{\% diff.} & 372 & 309 & 81 & 56.3 & -\\
\bottomrule
\end{tabular}

\label{LOComp.tab}
\end{table}


The presented results show that the learned observer is able to provide lower errors than the model-based observers. This is due to the ability of neural networks to adapt to the varying measurement noise and to estimate the vehicle dynamics. It can be seen in the longitudinal and lateral velocity estimations that the Dugoff-based and the Pacejka-based observers are able to have close errors to the learned observer for the low acceleration domain, while only the Pacejka-based observers have close errors to the learned observer in the high acceleration domain.

The presented results show the ability of learned techniques to overcome model validity issues, especially in the high acceleration scenarios where model-based methods present inaccuracies. The used learned observer was able to deliver more accurate estimates than the models with a Pacejka tire model, which is one of the highest fidelity models used in the literature. Thus, the interest in learning control applications rose in the recent years to overcome many problems including model validity and sensor noise reduction. The choice to use a learned model depends on the scenario encountered and the needed accuracy. 

\section{Conclusion}\label{conclusion.sec}
In this work, the issue of model validity was targeted and its effects on model-based control were investigated. 

We started by introducing several popular vehicle models used in the literature. We proceeded then to collect data from a real vehicle, to perform the accuracy analysis of the different models. Preliminary visualization tests were effected before moving to investigate the effect of lateral accelerations on the accuracy of each model. Two lateral acceleration domains split by the $a_y = 0.5g$ point were deduced and the behavior of the four models was drastically changing between the two domains. The validity of the models is concluded to be dependent on the lateral acceleration domain the vehicle is in. 

After proving the relationship between the acceleration domains and the accuracy of the models, the presented issues were implemented in a control-related application, a state observer. The effects of the model errors in high lateral acceleration domains were shown in observing applications. The model-based observers are not able to well perform whenever the model they use is not valid. 

After showing the effects of model validity on state observers, learning-based observers were briefly visited. A learning-based observer was implemented and its outperformance to the four model-based observers was proven, especially in the high lateral acceleration domain, where the validity of the used models is at risk. 

In brief, model validity is an important factor when implementing autonomous driving solutions, the used model should be able to represent the behavior of the vehicle to reach the desired accuracy. The choice of the model should be made based on the expected type of scenarios. This work showed the deficiency of simple models whenever a high acceleration domain is entered and the ability of high-fidelity models and learned models to produce lower errors in these scenarios. This emphasizes the importance of the choice of the model in planning and control applications for accurate and safe vehicle behavior.  

\section*{Acknowledgement}
We would like to thank Prof. Markus Maurer and Dr. Marcus Nolte at the Institute of Control Engineering, TU Braunschweig for enabling the collaboration that led to this work.

\bibliographystyle{ieeetr}
\bibliography{main}

\end{document}